\pdfoutput=1
\documentclass[11pt]{article}

\usepackage{ACL2023}
\usepackage{times}
\usepackage{latexsym}

\usepackage[T1]{fontenc}
\usepackage[utf8]{inputenc}

\usepackage{microtype}

\usepackage{inconsolata}
\usepackage{xcolor}
\usepackage{multirow}
\usepackage{booktabs}
\usepackage{makecell}
\usepackage{amssymb}
\usepackage{colortbl}
\usepackage{graphicx}
\usepackage{bbding}
\usepackage{utfsym}
\usepackage{caption}
\usepackage{subcaption}
\usepackage{amsmath}
\usepackage{enumitem}
\usepackage[normalem]{ulem}

\usepackage[export]{adjustbox}

\definecolor{darkgreen}{rgb}{0.0, 0.5, 0.0}
\usepackage{pifont}
\usepackage{xcolor}
\usepackage{xspace}
\definecolor{ForestGreen}{RGB}{34,139,34}
\definecolor{BrickRed}{rgb}{.72,0,0}
\definecolor{LakeBlue}{RGB}{0,61,153}

\newlength{\RoundedBoxWidth}
\newsavebox{\GrayRoundedBox}
   {\setlength{\RoundedBoxWidth}{\dimexpr#1}
    \begin{lrbox}{\GrayRoundedBox}
       \begin{minipage}{\RoundedBoxWidth}}%
   {   \end{minipage}
    \end{lrbox}
    \begin{center}
    \begin{tikzpicture}%
       \draw node[draw=black,fill=black!10,rounded corners,%
             inner sep=2ex,text width=\RoundedBoxWidth]%
             {\usebox{\GrayRoundedBox}};
    \end{tikzpicture}
    \end{center}}

\usepackage{tcolorbox}
\newtcolorbox{finding}{
colframe=black!80,
colback=icyrockblue,
fonttitle=\bfseries,
coltitle=black,
left=3pt,
right=3pt,
top=3pt,
bottom=3pt,
boxrule=0.4mm,
arc=3mm
}

\usepackage[utf8]{inputenc} % allow utf-8 input
\usepackage[T1]{fontenc}    % use 8-bit T1 fonts
\usepackage{hyperref}       % hyperlinks
\usepackage{url}            % simple URL typesetting
\usepackage{booktabs}       % professional-quality tables
\usepackage{amsfonts}       % blackboard math symbols
\usepackage{nicefrac}       % compact symbols for 1/2, etc.
\usepackage{microtype}      % microtypography
\usepackage{xcolor}         % colors
\usepackage{wrapfig}

\usepackage[ruled,boxed]{algorithm2e} 
\usepackage{algpseudocode}

\usepackage{inconsolata}
\usepackage{xcolor}
\usepackage{multirow}
\usepackage{booktabs}
\usepackage{makecell}
\usepackage{amssymb}
\usepackage{colortbl}
\usepackage{graphicx}
\usepackage{bbding}
\usepackage{utfsym}
\usepackage{caption}
\usepackage{subcaption}
\usepackage{pifont}
\usepackage{amsmath}
\usepackage{enumitem}
\usepackage[export]{adjustbox}% http://ctan.org/pkg/adjustbox
\usepackage{xspace}

\definecolor{ForestGreen}{RGB}{34,139,34}
\definecolor{BrickRed}{rgb}{.72,0,0}
\definecolor{LakeBlue}{RGB}{0,61,153}

\definecolor{self1}{RGB}{192,0,0}
\definecolor{self2}{RGB}{46,117,182}
\definecolor{self3}{RGB}{118,113,113}

\definecolor{ao(english)}{rgb}{0.0, 0.5, 0.0}
\definecolor{veronica-red}{RGB}{196,30,58}

\usepackage{fontawesome}
\definecolor{ForestGreen}{RGB}{34,139,34}
\definecolor{BrickRed}{rgb}{.72,0,0}
\definecolor{LakeBlue}{RGB}{0,61,153}

\newcommand{\ours}{\textsl{Genius}\xspace}

\title{\protect\includegraphics[scale=.035, valign=c]{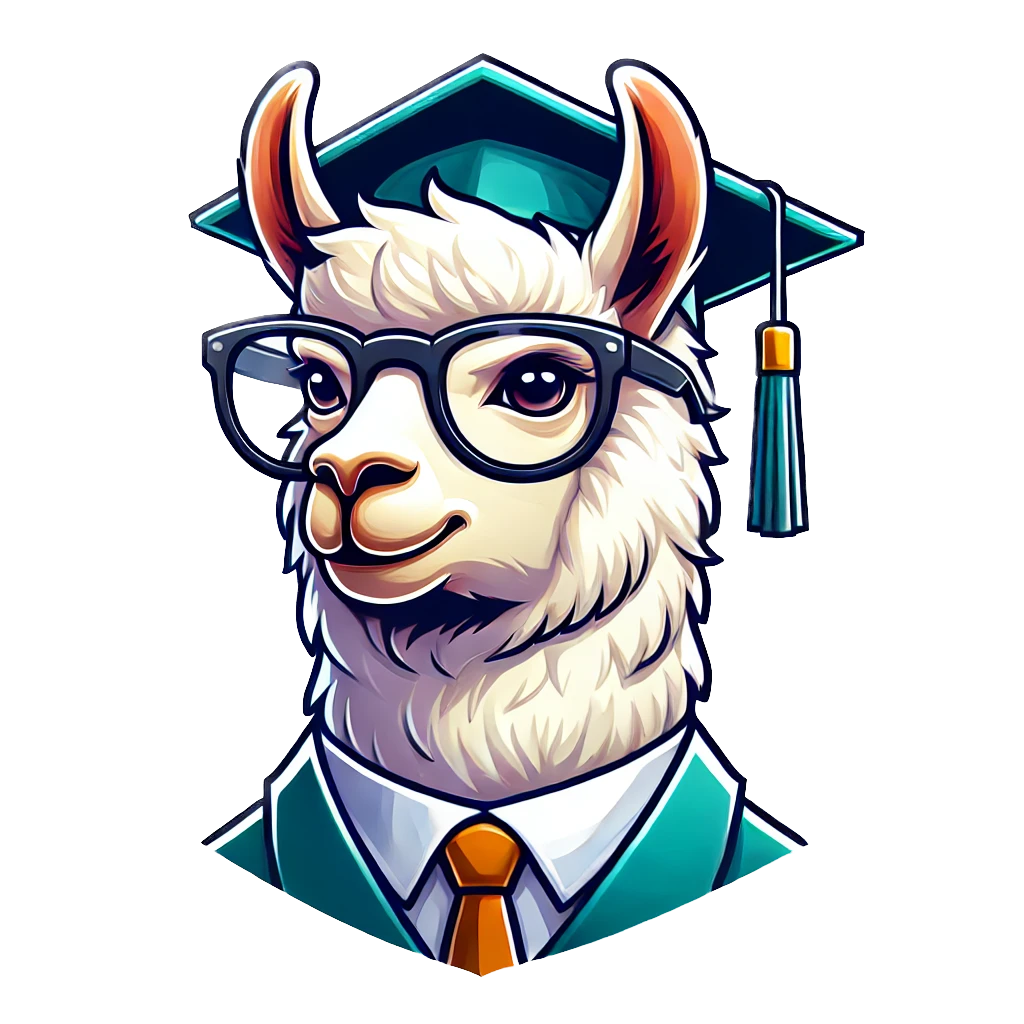}Genius: A \underline{Gen}eral\underline{i}zable and Purely Uns\underline{u}pervised \underline{S}elf-Training Framework For Advanced Reasoning}

\author{
Fangzhi Xu\textsuperscript{1,2}\thanks{\, Work done during internship at Shanghai AI Lab.} \quad
Hang Yan\textsuperscript{2} \quad
Chang Ma\textsuperscript{3} \quad
Haiteng Zhao\textsuperscript{4} \quad
Qiushi Sun\textsuperscript{1,3} \\
\bf{
Kanzhi Cheng\textsuperscript{1} \quad
Junxian He\textsuperscript{5} \quad
Jun Liu\textsuperscript{2}\footnotemark[2] \quad
Zhiyong Wu\textsuperscript{1}\thanks{\, Corresponding Author.} \quad
}\\
\textsuperscript{1}Shanghai AI Lab \quad
\textsuperscript{2}Xi'an Jiaotong University \quad
\textsuperscript{3}The University of Hong Kong \quad \\
\textsuperscript{4}Peking University \quad
\textsuperscript{5}Hong Kong University of Science and Technology \quad \\
\texttt{\{fangzhixu98, whucs2013wzy\}@gmail.com} \quad 
\texttt{liukeen@xjtu.edu.cn} \quad  \\
}

\begin{document}
\maketitle

\begin{abstract}
Advancing LLM reasoning skills has captivated wide interest.
However, current post-training techniques rely heavily on supervisory signals, such as outcome supervision or auxiliary reward models, which face the problem of scalability and high annotation costs.
This motivates us to \emph{enhance LLM reasoning without the need for external supervision}.
We introduce a \underline{gen}eral\underline{i}zable and purely uns\underline{u}pervised \underline{s}elf-training framework, named \ours.
Without external auxiliary, \ours requires to seek the optimal response sequence in a stepwise manner and optimize the LLM.
To explore the potential steps and exploit the optimal ones,
\ours introduces a stepwise foresight re-sampling strategy to sample and estimate the step value by simulating future outcomes.
Further, we recognize that the unsupervised setting inevitably induces the intrinsic noise and uncertainty.
To provide a robust optimization, 
we propose an advantage-calibrated optimization (ACO) loss function to mitigate estimation inconsistencies. 
Combining these techniques together,
\ours provides an advanced initial step towards self-improve LLM reasoning with general queries and without supervision,
revolutionizing reasoning scaling laws given the vast availability of general queries.
The code will be released at \url{https://github.com/xufangzhi/Genius}.

\end{abstract}
\section{Introduction}

\begin{figure}[t]
\large
\centering
\includegraphics[scale=0.7]{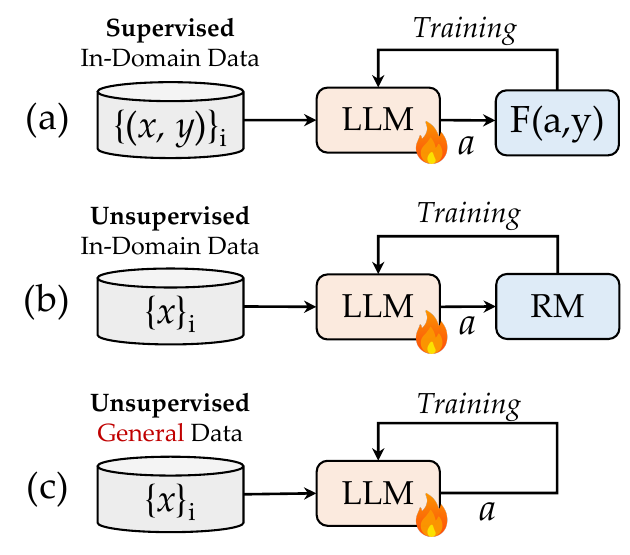}
\caption{Typical reinforce-like approaches to boost LLM reasoning. (a) and (b) abstract two types of reinforce-like methods, which require final answer and verification respectively. (c) depicts the ultimate goal of our research. $x$ denotes the input query, $a$ denotes the LLM-generated responses that contain multiple steps, and $y$ is the final answer (if exists).}
% \jh{what is ``a'' in this figure?}}
\label{intro}
\vspace{-1em}
\end{figure}

Reasoning skills are crucial for Large Language Models (LLMs) to achieve human-level intelligence~\cite{achiam2023gpt, team2023gemini}.
Recent efforts~\cite{qwq-32b-preview, guo2025deepseek} focus on enhancing reasoning capabilities during the post-training phase~\cite{li2025system}.
Typically, existing methods rely on supervision, which can be divided into two main approaches.
One stream is supervised fine-tuning (SFT) which requires the well-annotated response ($a$) paired with the query ($x$).
The other line of work is reinforce-like approaches that require either ground-truth answers or verification.
The former needs rule-based matching of the answer, as depicted in Fig.~\ref{intro}(a).
% One stream of methods (Fig.~\ref{intro}(a)) requires in-domain reasoning data with outcome supervision (e.g., mathematics) to give explicit answer verifications~\cite{trung2024reft,cheng2024vision}.
Though they are effective in specific domains like math and coding, many other problems lack clear solutions or explicit ground truth, presenting challenges for generalization to broader reasoning tasks.
% Therefore, extending these techniques to broader reasoning domains proves to be intractable.
% Another line of work expands the domain of reasoning tasks by (Fig.~\ref{intro}(b)) relying on the reward model to give supervision signals~\cite{jiao2024learning}.
The latter utilizes the external reward model for verification.
However, the training of a generalized reward model relies on expensive annotation~\citep{lightman2023let} and it may induce the reward hacking problem~\cite{amodei2016concrete, gao2023scaling}.
% These limitations largely affect the scalability of training reasoning models.
Considering these limitations of scalability,
it would revolutionize reasoning scaling law if we could achieve effective and efficient post-training simply from \emph{general queries}.
Therefore, it motivates us to raise a research question of \emph{\textbf{How to advance LLM reasoning ability without any external supervision ?}} (Fig.~\ref{intro}(c)).

This work tackles this problem by proposing a generalizable self-training framework, named \ours.
As depicted in Fig.~\ref{intro}(c), \ours only requires the policy LLM itself with a set of unsupervised queries.
Without the external auxiliary,
it builds upon the self-training paradigm that the LLM first generates response $a$ given the input query $x$, and selects the optimal ones for training.
As an initial attempt, \ours paves the way to self-improve LLM reasoning with unsupervised general queries.
% With merely 25K unsupervised general queries, \ours surprisingly improves the average performance of 7 reasoning benchmarks by >7\%. 

% \jh{self-training can have different types, better to use 1 or 2 sentences to clarify the self-training that we are talking about here, sth like it generates responses and selects responses to train.}

To generate training data for self-training, a crucial challenge is to determine \emph{how to collect and self-reward LLM responses without relying on external auxiliary resources.}
One intuitive solution is to have the LLM generate the entire response and then evaluate the response based on its sequence confidence~\cite{xu2024interactive}, re-prompt the model to assign scores~\cite{yuanself}, or apply self-consistency methods~\cite{prasad2024self}.
However, these attempts primarily focus on response-level rewards, while step-wise supervision could provide more stable and fine-grained training~\citep{uesato2022solving,lightman2023let,wang2024math}. Recent efforts like~\cite{zhang2024chain} employ step-level sampling, but they naturally inherit the short-sighted limitation of auto-regressive generation, lacking global adherence to overarching goals~\cite{ma2024non}. Search-based rewarding (e.g., MCTS-style methods) provides more global-awareness but requires intricate backtracking processes, which are notoriously time-consuming.

% To this end, \ours models the response as a sequence of steps and seeks the optimal ones via stepwise sampling and rewarding.
To this end, \ours seeks the optimal response sequence via stepwise sampling and collects high-quality preference pairs with rewards.
% \xfz{Motivated by xxx}
% It aims to seek the globally optimal response via stepwise sampling and rewarding.
% To determine the next step in the response, we first generate multiple candidate steps based on the preceding path.
Without reliance on external supervision to obtain the global-awareness, 
\ours adopts a coarse step estimation by simply rolling out future steps, referred to as \emph{foresight}~\cite{ma2024non,xu2025phi} and using their uncertainty as the foresight score to evaluate candidate steps.
Based on this technique,
we propose the stepwise foresight re-sampling approach:
using foresight scores to approximate the distribution that is sampled to determine the next step (for exploration) and re-sampled to create step-level preference pairs (for exploitation).

% To this end, we seek an efficient yet globally optimal annotation of process rewards.
% \ours employs a coarse estimation by simply rolling out future steps and using their outcomes as the score of the current step.
% \ours employs a stepwise foresight re-sampling approach: the step value is self-rewarded by simulating future outcomes of the current step. 
% The uncertainty in foresight paths then generates a distribution, which is sampled to guide subsequent steps (exploration) and re-sampled to create step-level preference pairs (exploitation).

Although the above approach offers a quality-efficiency balanced solution for superior \emph{sample-and-reward},
calculating the distribution of foresight scores based on a few rollouts may results in a biased estimation of step values, inevitably inducing noise to the self-supervision labels.
% Thus, it would inevitably induce noise under the unsupervised setting.
% That is, a higher level of foresight confidence does not necessarily result in a better response.
However, previous self-training methods simply employ either supervised fine-tuning~\cite{zelikman2022star} or reinforced learning strategies~\cite{yuanself} for optimization, neglecting the uncertainty of self-supervision~\cite{liang2024robust}. 
% While some works attempt to offer robust solutions~\cite{liang2024robust},
% these general optimization strategies overlook incorporating feedback from the sampling and self-rewarding process, leading to suboptimal outcomes.
We tackles this second challenge -- \emph{improving the robustness of self-training optimization} --
 by introducing an advantage-calibrated loss function (ACO), which penalizes inconsistent estimation between foresight score and step advantage. We find that ACO, compared to SFT~\citep{zelikman2022star}, and DPO~\citep{rafailov2024direct} improves training stability and boosts performance.

% Beyond the surprising experimental results,
\ours offers a unique perspective on post-training: LLMs can self-improve their general reasoning abilities using general queries without any form of external supervision. 
With merely 25K unsupervised general queries, \ours surprisingly improves the average performance across diverse reasoning benchmarks by >7\%. 
We also demonstrate that the scaling law on general queries consistently improves with more training steps (see \S~\ref{scale_law}). Given the abundance of general data available, this scaling can significantly enhance reasoning capabilities and further pushes the boundary of reasoning scaling law~\citep{guo2025deepseek}.
% We reveal that scaling law on general queries consistently improves with more training steps (refer to \S ~\ref{scale_law}).
% Given the abundance of general data available, the scaling would be very helpful for improving reasoning.

% Our findings illustrate a unique perspective on improving LLM reasoning abilities -- LLM itself contains the ability to enhance its reasoning ability without any forms of supervision. We reveal that scaling law on general data consistently improves with more training steps (cite results in \S 5.3), and considering the vastness of general data, the scaling would be very helpful for improving reasoning.

% \input{sections/1.introduction}
\section{Methodology}

\begin{figure*}[t]
\large
\centering
\includegraphics[scale=0.38]{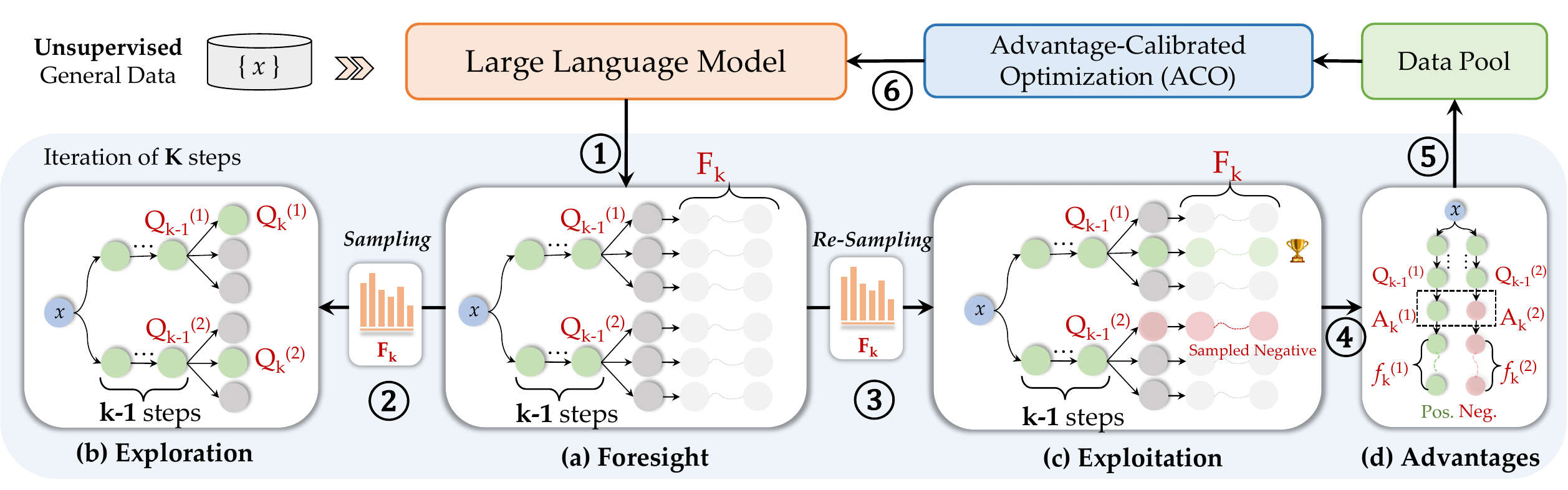}
\caption{The overall framework of \ours. It only receives the unsupervised NL queries as inputs. To complete goal of self-improving, the policy LLM goes through $K$ steps of sampling and rewarding for each query (Step 1-4), collects the high-quality response sequence as the training data (Step 5), and trains itself with the advantage calibrated optimization loss (Step 6).}
% \changh{The organization of the figure is a bit confusing. I was first reading left to right and then realized it's not in this order. Also the labels, e.g. step number, a/b/c/d should be larger.}}
% \sqs{more detailed caption required}
\label{model}
\vspace{-1em}
\end{figure*}

\subsection{Preliminary}
% Our framework, \ours, is presented in Figure~\ref{model}.
One major advantage of \ours is that it only requires unsupervised natural language (NL) queries as inputs. 
Under the self-training setting, the LLM $\pi_\theta$ generates responses given the query,
and then select the optimal ones to optimize itself. 
The main objectives of \ours are divided into two parts:
(1) synthesizing and rewarding the responses (\S~\ref{sec_foresight}); (2) optimizing the LLM with responses (\S~\ref{sec_aco}).
Figure~\ref{model} presents the overall framework of \ours.

To achieve the first objective,
\ours models the response as a sequence of steps.
The globally optimal response is derived by stepwise sampling and rewarding.
At each timestamp, \ours first rollouts a set of candidate steps and self-rewards them via simulating future steps (Fig.~\ref{model}(a)).
Then, we select the optimal next step (Fig.~\ref{model}(b)) and collect preference pairs for training (Fig.~\ref{model}(c)).
As for the second objective, \ours derives the step advantages during the sampling (Fig.~\ref{model}(d)), and adopts them to optimize the LLM with designed reinforced loss.

% In \ours, given the input query $x$, the policy LLM $\pi_\theta$ goes through the step-by-step exploration through foresight sampling [Step (a), (b)], 
% exploits the re-sampling to derive preference pairs with calculated step advantages [Step (c), (d)],
% and calibrate the optimization [Step (e)].

% \jh{this is an unclear paragraph to introduce the method. This paragraph should give a high-level description of the approach. For example, ``performs a maximum of $K$ steps exploration and exploitation'', not sure what this even means when reading this paragraph alone}
% In \ours, given each input query $x$,
% LLM policy $\pi_\theta$ performs a maximum of $K$ steps exploration and exploitation (\S~\ref{sec_foresight}).
% % At each step $k$, \ours collects training data to the pool.
% At each step $k$, \ours goes through foresight sampling\jh{the readers do not know what this is} (Fig.~\ref{model}(a)).
% Based on it, LLM first selects the optimal steps via sampling (Fig.~\ref{model}(b)) and curates the preference trajectories via re-sampling (Fig.~\ref{model}(c)).
% After obtaining abundant data from input queries, the LLM is optimized with the designed loss function (\S~\ref{sec_aco}).

For clearer illustration, we introduce the following symbols at the timestamp $k$:
$a_k$ denotes the current step with the step value of $Q_k$.
$\mathbf{a}_{<k}$ represents the preceding steps while $\mathbf{a'}_{>k}$ is the simulated future steps.
$T_k=(\mathbf{a}_{<k}, a_k, \mathbf{a'}_{>k})$ denotes the curated response for training.
For simplicity, we omit the symbol index in some descriptions.

% we use $a_k$ to denote the step at the timestamp $k$, and the corresponding step value is written as $Q_k$.

% At each step $k$, LLM rollouts a set of candidate steps $a_k$ based on preceding steps $\mathbf{a}_{<k}$ and then simulate the remaining steps $\mathbf{a'}_{>k}$ as foresight (Fig.~\ref{model}(a)).
% Based on it, LLMs select the steps via sampling on the uncertainty of foresight (Fig.~\ref{model}(b)), where we use $Q_k$ to denote the step value.

% At each step $k$, we use $Q^{(m)}_k$ to denote the value of step $k$ based on rollout beam $m$. Sampled path $T$ with k steps is denoted as $T = (a_1,a_2,\dots, a_k)$

% To curate training data, \ours select reasoning path $T_k$ at each timestamp $k$.

% In \ours, given each input query $x$,
%  LLM policy $\pi_\theta$ performs $K$-step exploration and exploitation via foresight re-sampling strategy to construct preference pairs (Step 1-5).
% After sampling the entire set of inputs,
% $\pi_\theta$ is optimized with the advantage-calibrated loss function (Step 6).
% % The overall implementation is concluded in the pseudocode in Appendix X.
% \changh{I think it would be better if you write the Q definition here. and also definition of T, just mention all the components you need for the training.} $Q^{(m)}_k$ denotes the value of step $k$ based on rollout beam $m$. Sampled path $T$ with k steps is denoted as $T = (a_1,a_2,\dots, a_k)$

\subsection{Exploration and Exploitation via Foresight Re-Sampling}
\label{sec_foresight}
% \jh{I feel the readers are still not very sure about the approach yet, and now we start talking about some details like diversity}
To ensure the diversity, we use beam search strategy~\citep{freitag2017beam} during the stepwise sampling.
We define the step beam size as $M$, as illustrated in Fig.~\ref{model} where we plot a simple case of $M=2$.

% For convenience, we take the $k^{th}$ step for illustration.

\paragraph{Step Rollouts with Foresight.}
At Step $k-1$, \ours keeps $M$ preceding paths $\mathbf{a}_{<k}$, each consisting of $k-1$ steps.
The value of the last step in the path is defined as $Q_{k-1}^{(m)}$, where $m \in [1,M]$.
For each beam $m$,
\ours first rollouts $N$ candidate steps $a_k$, leading to $M * N$ candidate steps totally.

To address the limitation of auto-regressive generation and construct the globally-aware response, \ours performs the simulation of future steps based on each candidate step $a_k$.
We refer to this process as \emph{foresight} (Fig.~\ref{model}(a)).
This allows us to derive the response sequence along with their respective foresight scores, calculated using the averaged log probability of the remaining steps:
% \begin{equation}
%     T,f_{k} \sim \pi_\theta(\cdot|\mathbf{a}_{<k};a_k).
% \end{equation}
\begin{equation}
    \mathbf{a}_{>k}',f_{k} \sim \pi_\theta(\cdot|\mathbf{a}_{<k};a_k).
\end{equation}
% For simplicity, we omit the index in the formula.
With these simulated future steps $\mathbf{a}_{>k}'$, the completed response can be constructed, written as $T=(\mathbf{a}_{<k},a_k,\mathbf{a}_{>k}')$.
% \changh{This score seems to only score k steps ?Does it include foresight?}
Here we obtain $M*N$ foresight scores $f_k$.
They can be utilized to approximate a distribution $\mathbf{F}_{k}$.
After normalization,
the elements of $\mathbf{F}_{k}$ are given by:
\begin{equation}
    \mathbf{F}_{k}(i) = \textrm{exp}( f_k^{(i)} / \tau) / \sum_{j}\textrm{exp}( f_k^{(j)} / \tau)
\end{equation}
In this expression, $\mathbf{F}_{k}(i)$ denotes the foresight score at index $i$, where $i \in [1,M*N]$. 
$\tau$ is the temperature parameter used to control the sharpness of the distribution.

% \changh{Distribution F is not defined ?Wonder if you need to keep both f and F}

\paragraph{Re-Sampling for Exploration and Exploitation.}

Based on the foresight technique,
\ours further selects the steps $a_{k}^{(m)}$ for current timestamp $k$ via sampling on the distribution of $\mathbf{F}_{k}$ (Fig.~\ref{model}(b)):
\begin{equation}
    \{a_k^{(m)}\}_{m=1}^{M} \sim \mathrm{Categorical}(\mathbf{F}_{k}).
\end{equation}
% The step value of $a_m^{(k)}$ is updated with the foresight score, 
% i.e., $Q_m^{(k)}:=f^{(k)}$. \changh{Perhaps also emphasize Q definition. Also, wonder if it‘dds necessary to define Q so early but only start using it now.}
In this way, we can keep $M$ beams for exploration in the next step.
Here, we define the $Q$ value of each selected step $a_k^{(m)}$ with the foresight score:
\begin{equation}
    Q_k^{(m)} := f_k^{(m)}
\end{equation}

Besides the exploration,
\ours also exploits the entire response sequence $T_k = (\mathbf{a}_{<k},a_k,\mathbf{a}_{>k}')$ at each timestamp $k$ for optimization (Fig.~\ref{model}(c)).
To encourage diversity and avoid overfitting on similar responses,
we introduce the re-sampling strategy based on the distribution $\mathbf{F}_{k}$.
The response with the highest foresight score $f_{k}^{w}$ is chosen as the positive one, written as $T_{k}^{w}$.
The negative response is re-sampled from $\mathbf{F}^{(k)}$:
\begin{equation}
    T_{k}^{l} \sim \mathrm{Categorical}(\mathbf{F}_{k} /\ f_{k}^{w})
\end{equation}
The corresponding foresight score of the negative path is $f_k^l$.
With such a re-sampling strategy, the balance between \emph{exploration} and \emph{exploitation} can be achieved.

\paragraph{Advantages and Data Construction.}
Since the reasoning sequences are completed from different beams,
it is insufficient to simply evaluate each step with the foresight score $f_{k}$.
Therefore, \ours derives the advantage value $A_{k}$ for both positive and negative response sequences:
\begin{equation}
    A_{k}^{w} = f_{k}^{w} - Q_{k-1}^{w}, \quad A_{k}^{l} = f_{k}^{l} - Q_{k-1}^{l}
\end{equation}
From the equation, the foresight score is calibrated with the $Q$ value of the previous step.

In this way, the training preference pair obtained from each step $k$ is constructed in the quintuple format, 
i.e., ($x$, $T_k^{w}$, $A_k^{w}$, $T_k^l$, $A_k^{l}$).

\subsection{Advantage-Calibrated Optimization}
\label{sec_aco}
% \changh{add some paragraphs in this section, so that have a general idea what each part of the equations is doing without carefully reading them. Just like 3.2}

Given the constructed preference pairs,
we can optimize the LLMs through reinforced learning.
There remains two critical steps unaddressed: (i) formulating the self-rewards for preference optimization; and (ii) deriving the optimization objective.

\paragraph{Formulating Self-Rewards as Preferences.}
Building upon Bradley-Terry model~\cite{Bradley1952RANKAO},
the measurement of the preferences can be formulated as:
\begin{equation}
    \label{eq_preference}
    \begin{split}
    p^*(T^w \succ T^l | x) &= \frac{\mathrm{exp}(r^*(x,T^w))}{\mathrm{exp}(r^*(x,T^l)) + \mathrm{exp}(r^*(x,T^l))} \\
    &= \sigma (r^*(x,T^w) - r^*(x,T^l))
    \end{split}
\end{equation}
where $r^*(T|x)$ represents the optimal reward function. $\sigma(\cdot) = 1 / (1 + \exp (-x))$ denotes the sigmoid function. 
% Based on this modeling, the reward model $r_\psi$ can be trained through $\min_{\psi} - \mathbb{E} [\mathrm{log} \sigma (r_{\psi}(T_w|x) - r_{\psi}(T_l|x) )]$. With the above reward model $r_\psi$, we can further optimize the policy LLM via RL fine-tuning:
% \begin{equation}
%     \max_{\theta} \mathbb{E}_{x \sim \mathcal{D}, T \sim \pi_\theta} [r_{\psi}(T|x)] \\ - \beta \mathrm{KL} (\pi_\theta(T|x) || \pi_\textrm{ref}(T|x)),
% \end{equation}
% where $\beta$ controls KL-divergence regularization.
Based on this modeling, the well-trained reward model $r_\psi$ is required to further optimize the LLM policy via RL fine-tuning.
However, under our unsupervised setting, training the reward model becomes impractical.

In the context of DPO~\cite{rafailov2024direct}, 
the policy LLM $\pi_\theta$ is leveraged as the implicit reward model. 
The self-reward function $\phi$ is modeled as:
\begin{equation}
    \begin{split}
    \phi(x,T) &= \beta \log \frac{\pi_\theta(T|x)}{\pi_\textrm{ref}(T|x)} + \beta \log Z(x) \\
    & \propto \beta \log \frac{\pi_\theta(T|x)}{\pi_\textrm{ref}(T|x)}
    \end{split}
\end{equation}
where $\pi_\textrm{ref}$ denotes the reference model and $Z(x) = \sum_{T} \pi_\textrm{ref}(T|x) \mathrm{exp} (\phi(x,T) / \beta)$ represents the partition function. 
With this approximation, 
the standard rewards for both the positive and negative response sequences can be derived:
\begin{equation}
    \phi_w = \beta \log \frac{\pi_\theta(T^w|x)}{\pi_\textrm{ref}(T^w|x)}, \, \phi_l = \beta \log \frac{\pi_\theta(T^l|x)}{\pi_\textrm{ref}(T^l|x)}
\end{equation}

\paragraph{ACO Loss Function.}

\begin{figure}[t]
\large
\centering
\includegraphics[scale=0.5]{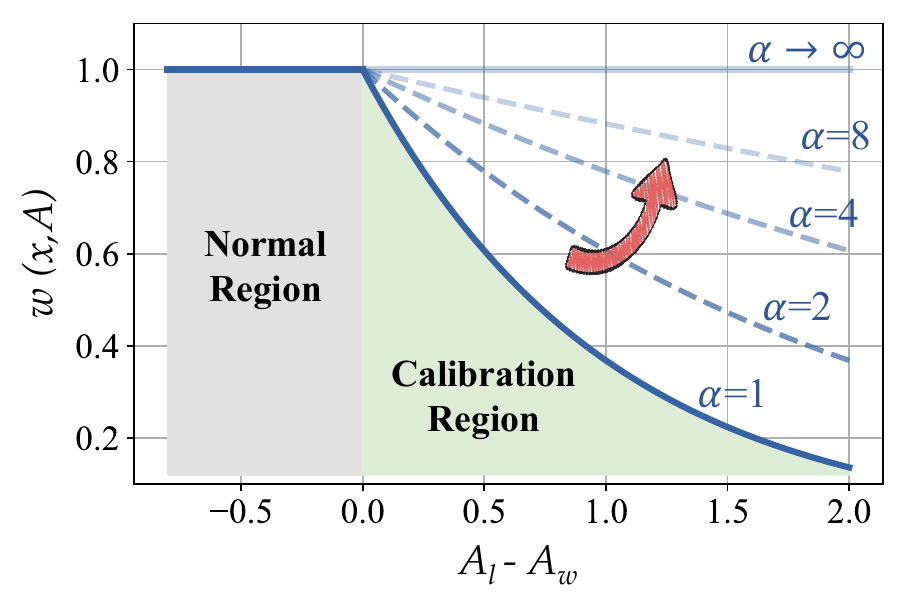}
\caption{Visualization of the calibration function. The x-axis denotes the the differences between $A_l$ and $A_w$, while the y-axis is the value of the calibration term. By adjusting $\alpha$, we can control the decay rate of the curve.}
\label{aco_loss}
% \vspace{-1em}
\end{figure}

% the standard DPO loss function can be formulated as:

% \begin{equation}
%     \phi(x,T) \propto \beta \frac{\pi_\theta(T|x)}{\pi_\textrm{ref}(T|x)}
% \end{equation}

% \begin{equation}
%     \mathcal{L}_\textrm{DPO} = - \mathbb{E}_{(x,T_w,T_l) \sim \mathcal{D}} \mathrm{log} \sigma (\phi(x,T_w) - \phi(x,T_l) )
% \end{equation}

Under the unsupervised setting, the training pairs are sampled based on the distribution of foresight score.
It would induce the noise during the optimization.
The above formulation of self-rewards treats each preference pair with equal scale,
making it difficult to detect the exception and improve the robustness.
% But it would be suboptimal under the unsupervised setting, where we only rely on the foresight value from the policy model itself to sample and derive the preferences.
% Therefore, we modify the selfreward for the negative sample and add the weighted advantage $w(x,T_l)$ as the relaxation term:
Therefore, we propose to employ the calculated advantage value $A$ to calibrate the self-reward function $\phi$.
In detail, we add the relaxation term $w(x,A)$ for the self-rewards of the negative response sequence:
\begin{equation}
    \phi_{l}(x,T^l) = \beta w(x, A) \log \frac{\pi_\theta(T^l|x)}{\pi_\textrm{ref}(T^l|x)},
\end{equation}
\begin{equation}
    w(x,A) = \mathrm{clip} \left( \exp \frac{- (A^{l}-A^{w})}{\alpha}, 1 \right)
\end{equation}
where $A_{l}-A_{w}$ denotes the differences in the advantages brought by the negative steps and positive steps. $\alpha$ is the hyper-parameter to control the scale of the relaxation term. 

For better understanding, we visualize $w(x,A)$ in Figure~\ref{aco_loss}.
This function can be categorized into two distinct regions: the \emph{Normal Region} and the \emph{Calibration Region}.
In the \emph{Normal Region}, where $A_{l}-A_{w} \leq 0$, the negative response sequence is distinguishable from the positive one.
Conversely, when $A_{l}-A_{w}>0$, the \emph{Calibration Region} is engaged, offering increased relaxation to the negative sample.
Through the adjustment of the hyper-parameter $\alpha$, we can regulate the extent of this flexibility.
In short, if the negative response sequence provides more actual advantages than the positive one, then it will be less punished (with smaller weight in the self-reward calculation).

% The modified loss function is as follow:
Substituting the self-reward function $\phi_w$ and $\phi_l$ into Eq.~\ref{eq_preference} and optimizing it using the form of negative log-likelihood,
the ACO loss is derived:
\begin{equation}
\begin{split}
    \mathcal{L}_\textrm{ACO} = - \mathbb{E}_{(x,T_w,T_l) \sim \mathcal{D}} \mathrm{log} \sigma \left[ \beta \log \frac{\pi_\theta(T_w|x)}{\pi_\textrm{ref}(T_w|x)} \right. \\
    - \left. \beta \mathrm{clip} \left( \exp \frac{- (A_{l}-A_{w})}{\alpha}, 1 \right) \log \frac{\pi_\theta(T_l|x)}{\pi_\textrm{ref}(T_l|x)} \right]
\end{split}
\end{equation}

We include the gradient analysis of ACO loss function and its relations to other robust preference optimization strategies in Appendix~\ref{app:gradient_derivation}.
% For ACO loss, the gradient is derived in Appendix~\ref{app:gradient_derivation}.

\section{Experiments}

\subsection{Implementation Details}

\paragraph{Training Corpora} 
The training queries are respectively sourced from two general corpora, Magpie~\cite{xu2024magpie} and OpenHermes-2.5~\cite{OpenHermes-2.5}.
Considering the computational cost, we opt to randomly select 25K queries from Magpie and 32K queries from OpenHermes-2.5.
They are utilized respectively as the sources for self-training.

\paragraph{Evaluation Tasks}
To comprehensively evaluate the basic reasoning abilities of the LLMs,
we incorporate the following benchmarks:
GSM8K~\cite{cobbe2021training}, MATH~\cite{hendrycks2021measuring}, and GPQA~\cite{rein2023gpqa} for math reasoning,
ReClor~\cite{yureclor} and LogiQA~\cite{liu2021logiqa} for logical reasoning,
StrategyQA~\cite{geva2021did} and ARC-Challenge~\cite{clark2018think} for general reasoning.
Moreover, we also include the some general benchmarks to verify the performance stability on the general domain: AlpacaEval~\cite{alpaca_eval}, WildBench~\cite{lin2024wildbench}, and ArenaHard~\cite{li2024crowdsourced} for subjective evaluation, WikiBench~\cite{kuo2024wikibench}, MMLU~\cite{hendrycks2020measuring}, and MMLU-Pro~\cite{wang2024mmlu} for objective evaluation.
Also, the competition-level benchmark AIME2024~\cite{aime_1983_2024} is included to prove the scalability.
% Please refer to Appendix~\ref{app:train_test} for more details.

% \subsection{Baselines and Base LLMs}

\paragraph{Baselines}
One line of baselines requires supervision (labeled response) beyond the training queries,
including \emph{SFT} and \emph{SPIN}~\cite{chenself}.
Another line of methods only needs unsupervised queries as the input,
covering \emph{STaR}~\cite{zelikman2022star}, \emph{CoH}~\cite{liuchain}, \emph{Self-Rewarding}~\cite{yuanself} and \emph{ScPO}~\cite{prasad2024self}. Details refer to Appendix~\ref{app:baselines}.

\paragraph{Base LLMs}
In the main experiments, we utilize LLaMA3.1-8B-Instruct~\cite{dubey2024llama} as the backbone.
To verify the generalization capability, we also apply the self-training approaches on Qwen2.5-Instruct series models~\cite{yang2024qwen2}, 
including 3B and 7B variants. 

\vspace{-0.1cm}
\paragraph{Training and Inference Setup}
For the foresight sampling configuration, we set $M$=2, $N$=4, and $K$=4.
Based on it, the total number of training pairs is 100K and 128K for Magpie and OpenHermes2.5 respectively.
% The LLM training is supported by Deepspeed Zero3 and FlashAttention2.
The inference process is accelerated by the vLLM engine.
Setup details refer to Appendix~\ref{app:setup}.

\subsection{Main Results}
\vspace{-0.1cm}
In Table~\ref{exp_main}, we present the evaluation results. 
In addition to presenting individual results for each dataset, the average performances are also reported in the last column.

\begin{table*}[t]
\centering
\footnotesize
\resizebox{\linewidth}{!}{
\begin{tabular}{l|ccccccc|c}
    \toprule
    % \multirow{1}{*}{\textbf{Models}} &\multicolumn{2}{c|}{\textbf{Math Reasoning}} &\multicolumn{2}{c|}{\textbf{Logical Reasoning}} &\multicolumn{4}{c}{\textbf{General Reasoning}} \\
    \multirow{1}{*}{\textbf{Models}} &\textbf{GSM8K} & \textbf{MATH} &\textbf{ReClor} &\textbf{LogiQA} &\textbf{StrategyQA} &\textbf{GPQA} &\textbf{ARC-c} &\textbf{Avg.} \\
    \midrule
    % LLaMA3.1-8B-Instruct \\
    LLaMA3.1-8B-Instruct (CoT) &70.28 &30.52 &49.40 &33.33 &58.91 &26.56 &78.33 &49.65 \\
    \midrule
    \multicolumn{9}{c}{\cellcolor{gray!25} Self-Training from \textbf{Magpie (25K)}}  \\
    \midrule
    \emph{w/ Supervision} & & & & & \\
    \quad SFT &71.72 &26.27 &52.80 &37.78 &57.34 &26.79 &74.06 &49.54  \\
    \quad SPIN~\cite{chenself} &74.91 &31.49 &\underline{57.40} &40.09 &\underline{71.35} &\underline{29.91} &\underline{83.96} &\underline{55.59} \\
    % \emph{w/ Self-reward Sampling} & & & & & \\
    % \quad + Traj. DPO &68.39 &27.20 &54.40 &36.71 \\
    % \quad + Step DPO &71.49 &26.60 &54.00 &36.10\\
    \midrule
    \emph{w/o Supervision} & & & & & \\
    % \quad ETO~\cite{song2024trial} \\
    \quad STaR~\cite{zelikman2022star} &72.86 &29.32 &46.40 &35.94 &33.36 &20.31 &67.24 &43.63 \\
    \quad CoH~\cite{liuchain} &74.37 &\underline{32.29} &56.20 &38.56 &69.08 &28.13 &82.51 &54.45 \\
    \quad Self-Rewarding~\cite{yuanself} &\underline{76.04} &30.19 &55.80 &37.94 &70.48 &28.35 &82.17 &54.42 \\
    \quad ScPO~\cite{prasad2024self} &71.11 &30.99 &55.00 &\underline{40.40} &59.87 &28.57 &78.92 &52.12 \\
    % \quad ENVISIONS~\cite{xu2024interactive} \\
    % \quad CPO~\cite{zhang2024chain} &ing & \\
    \midrule
    \quad \ours &\textbf{78.32} &\textbf{34.64} &\textbf{58.80} &\textbf{40.86} &\textbf{72.53} &\textbf{30.35} &\textbf{84.04} &\textbf{57.08} \\
    \midrule
    \multicolumn{9}{c}{\cellcolor{gray!25} Self-Training from \textbf{OpenHermes2.5 (32K)}}  \\
    \midrule
    \emph{w/ Supervision} & & & & & \\
    \quad SFT &63.68 &21.64 &45.00 &29.03 &48.47 &23.44 &69.37 &42.95 \\
    \quad SPIN~\cite{chenself} &63.61 &24.74 &54.00 &35.33 &59.00 &28.57 &71.76 &48.14 \\
    \midrule
    \emph{w/o Supervision} & & & & & \\
    % \quad ETO~\cite{song2024trial} &ing \\
    \quad STaR~\cite{zelikman2022star} &\underline{75.51} &29.47 &43.60 &34.87 &19.34 &22.99 &68.43 &42.03 \\
    \quad CoH~\cite{liuchain} &74.29 &31.22 &54.80 &38.40 &\underline{69.91} &29.69 &81.48 &\underline{54.26} \\
    \quad Self-Rewarding~\cite{yuanself} &73.92 &29.99 &\underline{56.00} &39.78 &67.55 &\underline{30.13} &\underline{81.66} &54.15 \\
    \quad ScPO~\cite{prasad2024self} &73.54 &\underline{31.27} &54.80 &\underline{41.01} &58.65 &28.79 &79.52 &52.51 \\
    % \quad ENVISIONS~\cite{xu2024interactive} \\
    % \quad CPO~\cite{zhang2024chain} &ing & \\
    % \emph{w/ Foresight Sampling} & & & & & \\
    \midrule
    \quad \ours &\textbf{75.82}	&\textbf{34.42} &\textbf{57.60}	&\textbf{41.63} &\textbf{70.79} &\textbf{34.82}	&\textbf{83.19} &\textbf{56.90} \\
    \bottomrule
\end{tabular}
}
\caption{Main Results. The self-training performances from Mappie and OpenHermes2.5 corpora are reported independently. The optimal results are in bold and the suboptimal ones are underlined.}
% \changh{Do we have a Q*/MCTS baseline just for one task perhaps. Reviewers may easily link your method to those methods after seeing the main figure.}}
\label{exp_main}
% \vspace{-0.5em}
\end{table*}

\vspace{-0.1cm}
\paragraph{\ours significantly boosts the reasoning abilities of LLaMA3.1, surpassing all strong baselines.}
With merely 25K unsupervised training queries (i.e., self-training from Magpie), \ours demonstrates a notable enhancement in the average CoT reasoning performance of LLaMA3.1-8B-Instruct by 7.43\%. 
This improvement is further underscored when using OpenHermes as the training corpus. 
In comparison to strong baselines, \ours consistently exhibits state-of-the-art performance, showcasing an average advantage of >2\%.
Among them, \ours presents obvious more advantages in challenging tasks (e.g., MATH),
outperforming \emph{Self-Rewarding} by >4\%.
% \hangyan{duplicated presentation}
Moreover, the superiority of \ours is consistent across all the evaluation benchmarks, while other baselines (e.g., CoH and SPIN) show deviations in performances.

% \changh{need to add more description of main results: detailed results of each benchmark, comparison to other baselines}

\paragraph{Self-training with RL optimization is more effective in improving reasoning with general data.}
Among the baselines, RL-based self-training methods (e.g., \emph{CoH}, \emph{Self-Rewarding} and \emph{ScPO}),
exhibit consistent advantages over supervised fine-tuning baselines (e.g., SFT and STaR).
Since current LLMs (e.g., LLaMA3.1) have been well optimized to perform chain-of-thought reasoning,
it requires RL to cultivate a broader generalization capacity, rather than relying on SFT to inject reasoning patterns.
Moreover, it is observed that the supervision of ground-truth responses does not yield enhancements, 
whereas self-generated responses can serve as an effective source to achieve performance boosts.

\paragraph{Performance Consistency on General Tasks.}
% Table~\ref{exp_general} reports the 
Beyond the reasoning-intensive tasks, it is also non-trivial to maintain the performances on the general benchmarks after post-training.
Supported by the evaluation suite of OpenCompass~\cite{contributors2023opencompass,cao2024compassjudger}, we experiment on 6 widely-used general benchmarks, which are categorized into \emph{subjective} and \emph{objective} evaluation.
We report the evaluation results in Table~\ref{exp_general}.

Overall, \ours keeps the stability of the performances on the general domain, with slight improvements in most cases.
Specifically, self-training with \ours also achieves huge performance gains in Arena-Hard benchmark, which reflects the superior alignment with human preference.
On the knowledge-intensive benchmarks (i.e., WikiBench, MMLU, and MMLU-Pro), \ours can maintain the performances of the base LLMs, avoiding catastrophic forgetting after post-training.

\subsection{Generalization and Adaptation}
\begin{figure}[t]
  \centering
  \begin{subfigure}{0.48\textwidth}
    \centering
    \includegraphics[width=\linewidth]{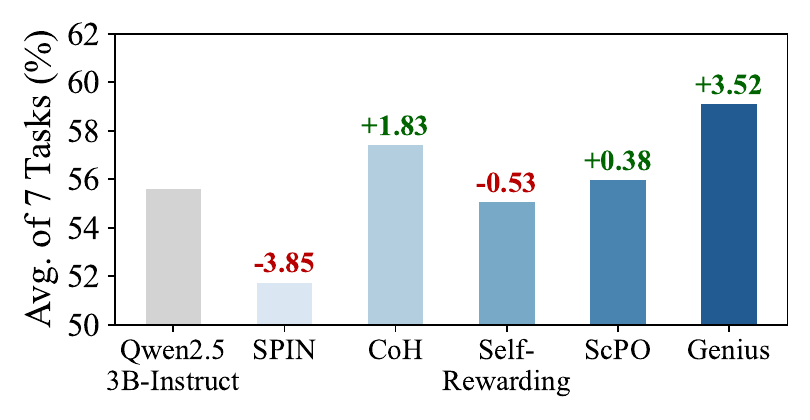}
    \vspace{-0.7cm}
    \caption{Base LLM: Qwen2.5-3B-Instruct}
    \label{fig:qwen-3b}
  \end{subfigure}
  \\
  \begin{subfigure}{0.48\textwidth}
    \centering
    \includegraphics[width=\linewidth]{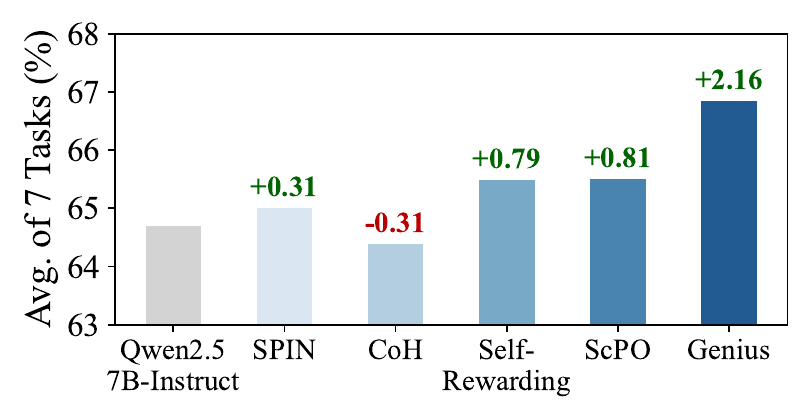}
    \vspace{-0.7cm}
    \caption{Base LLM: Qwen2.5-7B-Instruct}
    \label{fig:qwen-7b}
  \end{subfigure}
  \caption{Generalization to Qwen2.5 series models. All methods are trained on the OpenHermes2.5 split. The numbers above the bars represent the average performance gain relative to the base model.}
  % \changh{Maybe use a brighter color just for Genius}}
  \label{fig:qwen}
\end{figure}

% \changh{Need to rethink the title for this section.}

In this section, we analyze the generalization and adaptation of \ours to (i) different backbone LLMs, and (ii) on the competition-level task.

\vspace{-0.1cm}
\paragraph{Generalization to other backbone LLMs.}

Besides the experiments on LLaMA3.1, we also supplement the evaluations on Qwen2.5-series.
Fig.~\ref{fig:qwen-3b} and~\ref{fig:qwen-7b} present the average performance across 7 benchmarks on Qwen2.5-3B-Instruct and Qwen2.5-7B-Instruct respectively.
Compared with all the strong baselines,
\ours shows the highest performance gains over the base LLM.
Notably, self-training on Qwen2.5 series models does not yield larger benefits than on LLaMA3.1-8B-Instruct.
Some of the baselines even fail in some cases.
One hypothesis is that Qwen2.5-Instruct has conducted comprehensive post-training,
which makes it challenging for further advancement.
It does not conflict with our key contribution that \ours serves as a versatile post-training technique,
as it can function both as an ongoing self-training method for post-trained LLMs and as an alternative post-training strategy for the model itself.

\begin{table*}[t]
\centering
\footnotesize
\resizebox{\linewidth}{!}{
\begin{tabular}{l|ccc|ccc}
    \toprule
    % \multirow{1}{*}{\textbf{Models}} &\multicolumn{2}{c|}{\textbf{Math Reasoning}} &\multicolumn{2}{c|}{\textbf{Logical Reasoning}} &\multicolumn{4}{c}{\textbf{General Reasoning}} \\
    \multirow{2}{*}{\textbf{Models}} &\multicolumn{3}{c|}{\textbf{Subjective Benchmarks}} &\multicolumn{3}{c}{\textbf{Objective Benchmarks}} \\
    &\textbf{AlpacaEval} &\textbf{WildBench} &\textbf{Arena-H} &\textbf{WikiBench} &\textbf{MMLU} &\textbf{MMLU-Pro} \\
    \midrule
    LLaMA3.1-8B-Instruct &24.60 &-1.11 &30.31 &27.65 &71.14 &48.62 \\
    \textbf{\ours} [From Magpie (25K)] &\textbf{26.96} &\textbf{2.68} &\textbf{50.00} &\textbf{28.75} &71.86 &48.44 \\
    \textbf{\ours} [From OpenHermes (32K)] &25.47 &1.44 &\textbf{50.00} &27.00 &\textbf{72.21} &\textbf{49.19} \\
    \bottomrule
\end{tabular}
}
\caption{Performances on benchmarks in the general domain.}
\label{exp_general}
% \vspace{-0.5em}
\end{table*}

\paragraph{Adaptation to Challenging Task.}

\begin{figure}[t]
\large
\centering
\includegraphics[scale=0.58]{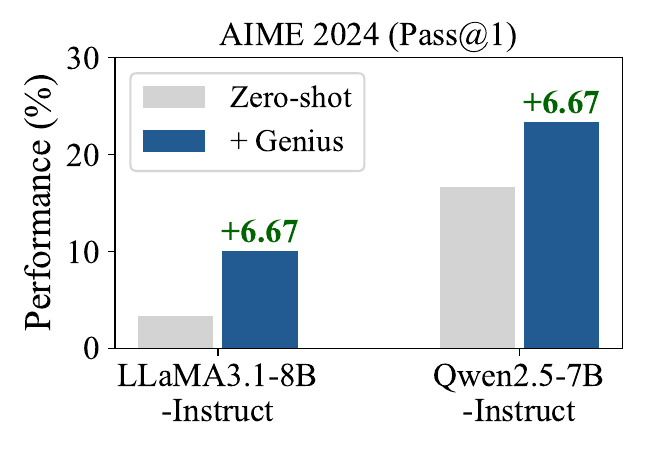}
\vspace{-0.3cm}
\caption{Results on AIME 2024.}
\label{aime}
% \vspace{-1em}
\end{figure}

Although \ours is not targeted for training large reasoning model like DeepSeek-R1, it is interesting to evaluate the challenging problems and uncover the potential.
Figure~\ref{aime} supplements the experiments on the competition-level task AIME 2024.
We evaluate the model trained on OpenHermes2.5 split from LLaMA3.1-8B-Instruct and Qwen2.5-7B-Instruct respectively.
It is observed that \ours boosts the performances by 6.67\%.
It verifies that \ours not only improves upon the fundamental LLM reasoning capabilities but also expands the boundaries of LLMs to address more intricate scenarios.
\section{Analysis}

\subsection{Ablation Studies}

To uncover the effectiveness of the major contributions of \ours,
\emph{sample-and-reward} strategy and \emph{optimization} objectives are ablated respectively.

\paragraph{Ablation of Sampling Strategy.}

\begin{table*}[t]
\centering
\footnotesize
\resizebox{\linewidth}{!}{
\begin{tabular}{l|ccccccc|cc}
    \toprule
    \multirow{1}{*}{\textbf{Variants}} &\textbf{GSM8K} & \textbf{MATH} &\textbf{ReClor} &\textbf{LogiQA} &\textbf{StrategyQA} &\textbf{GPQA} &\textbf{ARC-c} &\textbf{Avg.} &$\Delta$ \\
    \midrule
    \multicolumn{10}{c}{\cellcolor{gray!25} Self-Training from \textbf{Magpie (25K)}}  \\
    \midrule
    \ours &\textbf{78.32} &\textbf{34.64} &\textbf{58.80} &\textbf{40.86} &\textbf{72.53} &\textbf{30.35} &\textbf{84.04} &\textbf{57.08} &- \\
    \quad w/o \emph{foresight} &71.65 &31.07 &57.60 &39.17 &66.20 &30.13 &81.57 &53.91 &{\color{darkgreen} +3.17} \\
    \quad w/o \emph{sampling} &69.29 &31.37 &57.60 &39.17 &68.03 &24.33 &81.06 &52.98 &{\color{darkgreen} +4.10} \\
    \midrule
    \multicolumn{10}{c}{\cellcolor{gray!25} Self-Training from \textbf{OpenHermes2.5 (32K)}}  \\
    \midrule
    \ours &\textbf{75.82}	&\textbf{34.42} &\textbf{57.60}	&\textbf{41.63} &\textbf{70.79} &\textbf{34.82}	&\textbf{83.19} &\textbf{56.90} &- \\
    \quad w/o \emph{foresight} &73.01 &30.74 &57.60 &35.94 &67.25 &29.46 &81.57 &53.65 &{\color{darkgreen} +3.25}\\
    \quad w/o \emph{sampling} &72.93 &30.59 &56.00 &41.17 &65.76 &30.13 &80.03 &53.80 &{\color{darkgreen} +3.10} \\
    \bottomrule
\end{tabular}
}
\caption{Ablation studies on the sampling strategy. \emph{w/o foresight} ablates the look-ahead process and simply samples from the distribution formed by step uncertainty. \emph{w/o sampling} indicates that we adopt a greedy approach by selecting the two steps with the highest foresight score for \emph{exploration}, while the step with the lowest foresight score serves as the negative response for \emph{exploitation}.}
\label{exp_ablate}
\vspace{-0.9em}
\end{table*}

Table~\ref{exp_ablate} presents the ablation results. 
Firstly, ablating the foresight module results in 3.17\%-3.25\% average performance drops.
It illustrates that the foresight sampling strategy alleviates the short-sightedness of language model generation,
and the employment of foresight score optimizes the self-rewarding of the step value.
Secondly, replacing the \emph{sampling} with the greedy selection also leads to significant drops.
It verifies that our \emph{re-sampling} strategy strikes the balance between exploration and exploitation.

\paragraph{Ablation of Optimization Methods.}

\begin{table}[t]
\centering
\footnotesize
\resizebox{\linewidth}{!}{
\begin{tabular}{l|cc}
    \toprule
    \multirow{2}{*}{\textbf{Models}} &\multicolumn{2}{c}{\textbf{Average Performances}} \\
    & Magpie & OpenHermes \\
    \midrule
    LLaMA3.1-8B-Instruct (CoT) &49.65 &49.65 \\
    \midrule
    \multicolumn{3}{c}{\cellcolor{gray!25} w/ Foresight Sampling}  \\
    \midrule
    + \textbf{ACO} &\textbf{57.08} &\textbf{56.90} \\
    + DPO~\cite{rafailov2024direct} &55.51 &55.73 \\
    + SimPO~\cite{meng2025simpo} &50.42 &50.87 \\
    + IPO~\cite{azar2024general} &52.31 &52.20 \\
    + ROPO~\cite{liang2024robust} &55.30 &55.25 \\
    + SFT &44.63 &49.70 \\
    % \midrule
    % Mistral-Ins-v0.3 (7B) & &\\
    % + \emph{Foresight Sampling} & \\
    % \quad + \textbf{AW-DPO} \\
    % \quad + DPO \\
    % \quad + SimPO \\
    % \quad + ROPO \\
    % \quad + cDPO \\
    % \quad + IPO \\
    \bottomrule
\end{tabular}
}
\caption{Comparison of different reinforced loss. In the experiments, we only replace the optimization methods while keep the foresight sampling strategy unchanged.}
\vspace{-0.2cm}
\label{exp_ablate_dpo}
\end{table}

We present the comparison between various optimization approaches in Table~\ref{exp_ablate_dpo},
covering DPO, SimPO, IPO, ROPO and SFT.
Among these popular approaches, our ACO loss function stands out, showcasing significant advantages with an average performance improvement across 7 reasoning benchmarks.
Compared to the robustness optimization strategy ROPO, 
ACO is more suitable for self-training scenarios.

\subsection{Post-Training Scaling Law}
\label{scale_law}

\begin{figure}[t]
\large
\centering
\includegraphics[scale=0.53]{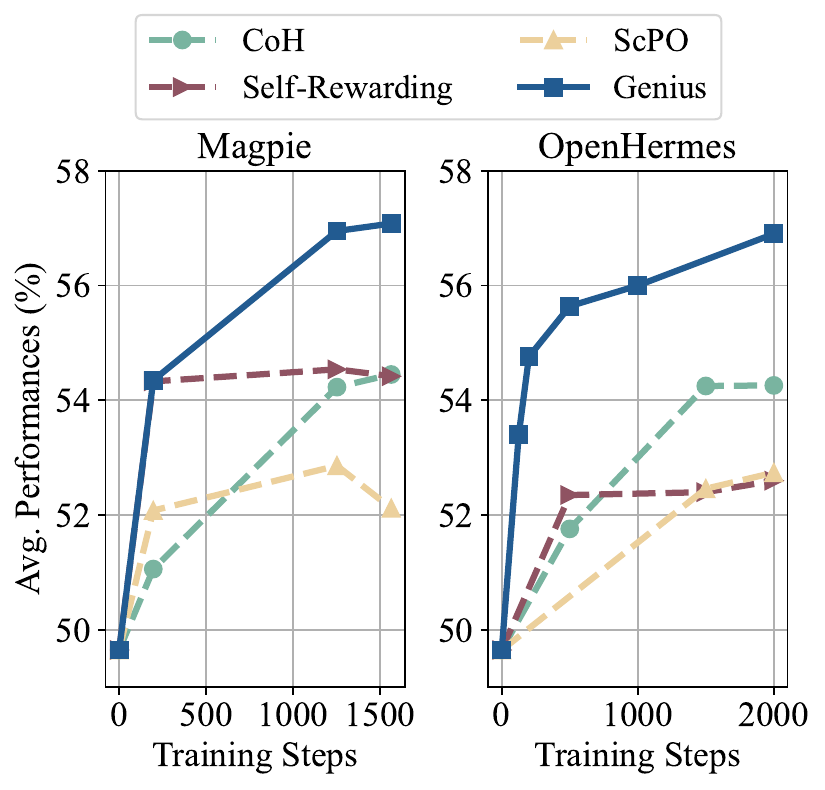}
\caption{Post-training scaling law with LLaMA3.1-8B-Instruct as the base LLM.}
\vspace{-0.2cm}
% \changh{the colors are not distinguishable enough. Use very different colors would be better.} }
\label{scaling}
% \vspace{-1em}
\end{figure}

Limited by computational resources,
we only provide the 10K-level training attempts, presented in the main table.
To give the direction of future post-training scaling,
Figure~\ref{scaling} shows the down-sampled scaling curves.
We have the following findings.

\paragraph{\ours holds great potential for further scalability.}
From Figure~\ref{scaling},
it is observed that \ours can rapidly self-improve with limited training steps.
The evolution progress proceeds smoothly with the training steps increasing.
This curve suggests that self-training with \ours is far from saturation and still has room for improvement,
whereas other baseline methods appear to face challenges when expanded in scale.

\vspace{-0.2cm}
\section{Related Works}
\paragraph{Post-Training for LLM Reasoning.} 
% \chang{Might emphasize on specifically, scaling up reasoning training in post-training stage. }
Boosting LLM reasoning by post-training has been a hot topic recently.
Some efforts perform large-scale imitation learning on well-curated reasoning data~\cite{yuemammoth, toshniwal2024openmathinstruct, xu2025redstar} to build large reasoning models for the specific domain (e.g., math, code).
Considering the huge annotation costs, some recent endeavors have turned to self-training techniques to synthesize the reasoning trajectories more efficiently~\cite{sun2024genesis}.
However, these efforts still rely on supervision signals and in-domain training corpus, either with explicit outcome supervision~\cite{zelikman2022star, xu2024interactive, trung2024reft, cheng2024vision} or auxiliary reward models~\cite{zeng2024b, jiao2024learning}. 
Conversely, we embark on a new path towards self-improving the general reasoning ability of LLMs without any form of supervision.

\paragraph{Optimization Techniques.} 
% \chang{-> Improving LLM without Supervision, feels like using self-play alot undermines novelty}
Previous works~\cite{yuemammoth, hu2023instructcoder, wang2025critique} often rely on supervised fine-tuning to inject new reasoning patterns (e.g., reflection, refinement) into the foundational LLMs.
With the development of stronger backbones,
some efforts~\cite{prasad2024self, cui2025processreinforcementimplicitrewards, guo2025deepseek} focus on offering the RL-based solution~\cite{rafailov2024direct, shao2024deepseekmath} to improve LLM reasoning,
proving the RL scaling effect with outcome rewards.
In our work, we propose a novel reinforced learning objective to perform robust optimization under the unsupervised setting.
We uncover the promising scaling law of self-training with RL optimization.

\section{Conclusion}
\vspace{-0.2cm}
This paper focuses on tackling the challenging and critical task of enhancing LLM reasoning, without relying on any external supervision.
A generalizable and purely unsupervised self-training framework, named \ours, is proposed to address several key technical challenges on how to: (1) sample responses; (2) self-reward responses without external auxiliary; (3) robustly optimize with the self-curated data.
Extensive experiments on 7 reasoning benchmarks, 7 general benchmarks, and 1 competition-level math task are included to comprehensively evaluate the LLM performance after post-training.
The analysis of the scaling law curve uncovers the huge potential on further scalability.

% \section*{Limitations}

% Limited by the computational resources, we only evaluate on 3B, 7B and 8B-sized LLMs, lacking the exploration of larger backbone LLMs.
% Also, the number of training queries is 25K and 32K respectively.
% Given that the scaling curve continues to show a trend of improvement, scaling up to larger input data sizes becomes an intriguing prospect.

\bibliography{custom}

\begin{thebibliography}{62}
\expandafter\ifx\csname natexlab\endcsname\relax\def\natexlab#1{#1}\fi

\bibitem[{Achiam et~al.(2023)Achiam, Adler, Agarwal, Ahmad, Akkaya, Aleman, Almeida, Altenschmidt, Altman, Anadkat et~al.}]{achiam2023gpt}
Josh Achiam, Steven Adler, Sandhini Agarwal, Lama Ahmad, Ilge Akkaya, Florencia~Leoni Aleman, Diogo Almeida, Janko Altenschmidt, Sam Altman, Shyamal Anadkat, et~al. 2023.
\newblock \href {https://arxiv.org/abs/2303.08774} {Gpt-4 technical report}.
\newblock \emph{arXiv preprint arXiv:2303.08774}.

\bibitem[{Amodei et~al.(2016)Amodei, Olah, Steinhardt, Christiano, Schulman, and Man{\'e}}]{amodei2016concrete}
Dario Amodei, Chris Olah, Jacob Steinhardt, Paul Christiano, John Schulman, and Dan Man{\'e}. 2016.
\newblock \href {https://arxiv.org/abs/1606.06565} {Concrete problems in ai safety}.
\newblock \emph{arXiv preprint arXiv:1606.06565}.

\bibitem[{Austin et~al.(2021)Austin, Odena, Nye, Bosma, Michalewski, Dohan, Jiang, Cai, Terry, Le et~al.}]{austin2021program}
Jacob Austin, Augustus Odena, Maxwell Nye, Maarten Bosma, Henryk Michalewski, David Dohan, Ellen Jiang, Carrie Cai, Michael Terry, Quoc Le, et~al. 2021.
\newblock \href {https://arxiv.org/abs/2108.07732} {Program synthesis with large language models}.
\newblock \emph{arXiv preprint arXiv:2108.07732}.

\bibitem[{Azar et~al.(2024)Azar, Guo, Piot, Munos, Rowland, Valko, and Calandriello}]{azar2024general}
Mohammad~Gheshlaghi Azar, Zhaohan~Daniel Guo, Bilal Piot, Remi Munos, Mark Rowland, Michal Valko, and Daniele Calandriello. 2024.
\newblock \href {https://proceedings.mlr.press/v238/gheshlaghi-azar24a.html} {A general theoretical paradigm to understand learning from human preferences}.
\newblock In \emph{International Conference on Artificial Intelligence and Statistics}, pages 4447--4455. PMLR.

\bibitem[{Bradley and Terry(1952)}]{Bradley1952RANKAO}
Ralph~Allan Bradley and Milton~E. Terry. 1952.
\newblock \href {https://api.semanticscholar.org/CorpusID:121987403} {Rank analysis of incomplete block designs the method of paired comparisons}.
\newblock \emph{Biometrika}, 39:324--345.

\bibitem[{Cao et~al.(2024)Cao, Lam, Duan, Liu, Zhang, and Chen}]{cao2024compassjudger}
Maosong Cao, Alexander Lam, Haodong Duan, Hongwei Liu, Songyang Zhang, and Kai Chen. 2024.
\newblock \href {https://arxiv.org/abs/2410.16256} {Compassjudger-1: All-in-one judge model helps model evaluation and evolution}.
\newblock \emph{arXiv preprint arXiv:2410.16256}.

\bibitem[{Chen et~al.(2024)Chen, Deng, Yuan, Ji, and Gu}]{chenself}
Zixiang Chen, Yihe Deng, Huizhuo Yuan, Kaixuan Ji, and Quanquan Gu. 2024.
\newblock \href {https://arxiv.org/abs/2401.01335} {Self-play fine-tuning converts weak language models to strong language models}.
\newblock In \emph{Forty-first International Conference on Machine Learning}.

\bibitem[{Cheng et~al.(2024)Cheng, Li, Xu, Zhang, Zhou, and Liu}]{cheng2024vision}
Kanzhi Cheng, Yantao Li, Fangzhi Xu, Jianbing Zhang, Hao Zhou, and Yang Liu. 2024.
\newblock \href {https://arxiv.org/abs/2411.00855} {Vision-language models can self-improve reasoning via reflection}.
\newblock \emph{arXiv preprint arXiv:2411.00855}.

\bibitem[{Clark et~al.(2018)Clark, Cowhey, Etzioni, Khot, Sabharwal, Schoenick, and Tafjord}]{clark2018think}
Peter Clark, Isaac Cowhey, Oren Etzioni, Tushar Khot, Ashish Sabharwal, Carissa Schoenick, and Oyvind Tafjord. 2018.
\newblock \href {https://arxiv.org/abs/1803.05457} {Think you have solved question answering? try arc, the ai2 reasoning challenge}.
\newblock \emph{arXiv preprint arXiv:1803.05457}.

\bibitem[{Cobbe et~al.(2021)Cobbe, Kosaraju, Bavarian, Chen, Jun, Kaiser, Plappert, Tworek, Hilton, Nakano et~al.}]{cobbe2021training}
Karl Cobbe, Vineet Kosaraju, Mohammad Bavarian, Mark Chen, Heewoo Jun, Lukasz Kaiser, Matthias Plappert, Jerry Tworek, Jacob Hilton, Reiichiro Nakano, et~al. 2021.
\newblock \href {https://arxiv.org/abs/2110.14168} {Training verifiers to solve math word problems}.
\newblock \emph{arXiv preprint arXiv:2110.14168}.

\bibitem[{Contributors(2023)}]{contributors2023opencompass}
OpenCompass Contributors. 2023.
\newblock \href {https://github.com/open-compass/opencompass} {Opencompass: A universal evaluation platform for foundation models}.
\newblock \emph{GitHub repository}.

\bibitem[{Cui et~al.(2025)Cui, Yuan, Wang, Wang, Li, He, Fan, Yu, Xu, Chen, Yuan, Chen, Zhang, Lv, Wang, Yao, Han, Peng, Cheng, Liu, Sun, Zhou, and Ding}]{cui2025processreinforcementimplicitrewards}
Ganqu Cui, Lifan Yuan, Zefan Wang, Hanbin Wang, Wendi Li, Bingxiang He, Yuchen Fan, Tianyu Yu, Qixin Xu, Weize Chen, Jiarui Yuan, Huayu Chen, Kaiyan Zhang, Xingtai Lv, Shuo Wang, Yuan Yao, Xu~Han, Hao Peng, Yu~Cheng, Zhiyuan Liu, Maosong Sun, Bowen Zhou, and Ning Ding. 2025.
\newblock \href {http://arxiv.org/abs/2502.01456} {Process reinforcement through implicit rewards}.

\bibitem[{Dubey et~al.(2024)Dubey, Jauhri, Pandey, Kadian, Al-Dahle, Letman, Mathur, Schelten, Yang, Fan et~al.}]{dubey2024llama}
Abhimanyu Dubey, Abhinav Jauhri, Abhinav Pandey, Abhishek Kadian, Ahmad Al-Dahle, Aiesha Letman, Akhil Mathur, Alan Schelten, Amy Yang, Angela Fan, et~al. 2024.
\newblock \href {https://arxiv.org/abs/2407.21783} {The llama 3 herd of models}.
\newblock \emph{arXiv preprint arXiv:2407.21783}.

\bibitem[{Freitag and Al-Onaizan(2017)}]{freitag2017beam}
Markus Freitag and Yaser Al-Onaizan. 2017.
\newblock \href {https://arxiv.org/abs/1702.01806} {Beam search strategies for neural machine translation}.
\newblock \emph{arXiv preprint arXiv:1702.01806}.

\bibitem[{Furuta et~al.(2025)Furuta, Lee, Gu, Matsuo, Faust, Zen, and Gur}]{furuta2025geometric}
Hiroki Furuta, Kuang-Huei Lee, Shixiang~Shane Gu, Yutaka Matsuo, Aleksandra Faust, Heiga Zen, and Izzeddin Gur. 2025.
\newblock \href {https://arxiv.org/pdf/2409.06691} {Geometric-averaged preference optimization for soft preference labels}.
\newblock \emph{Advances in Neural Information Processing Systems}, 37:57076--57114.

\bibitem[{Gao et~al.(2023)Gao, Schulman, and Hilton}]{gao2023scaling}
Leo Gao, John Schulman, and Jacob Hilton. 2023.
\newblock \href {https://proceedings.mlr.press/v202/gao23h.html} {Scaling laws for reward model overoptimization}.
\newblock In \emph{International Conference on Machine Learning}, pages 10835--10866. PMLR.

\bibitem[{Geva et~al.(2021)Geva, Khashabi, Segal, Khot, Roth, and Berant}]{geva2021did}
Mor Geva, Daniel Khashabi, Elad Segal, Tushar Khot, Dan Roth, and Jonathan Berant. 2021.
\newblock \href {https://direct.mit.edu/tacl/article/doi/10.1162/tacl_a_00370/100680/Did-Aristotle-Use-a-Laptop-A-Question-Answering} {Did aristotle use a laptop? a question answering benchmark with implicit reasoning strategies}.
\newblock \emph{Transactions of the Association for Computational Linguistics}, 9:346--361.

\bibitem[{Guo et~al.(2025)Guo, Yang, Zhang, Song, Zhang, Xu, Zhu, Ma, Wang, Bi et~al.}]{guo2025deepseek}
Daya Guo, Dejian Yang, Haowei Zhang, Junxiao Song, Ruoyu Zhang, Runxin Xu, Qihao Zhu, Shirong Ma, Peiyi Wang, Xiao Bi, et~al. 2025.
\newblock \href {https://arxiv.org/abs/2501.12948} {Deepseek-r1: Incentivizing reasoning capability in llms via reinforcement learning}.
\newblock \emph{arXiv preprint arXiv:2501.12948}.

\bibitem[{Hendrycks et~al.(2020)Hendrycks, Burns, Basart, Zou, Mazeika, Song, and Steinhardt}]{hendrycks2020measuring}
Dan Hendrycks, Collin Burns, Steven Basart, Andy Zou, Mantas Mazeika, Dawn Song, and Jacob Steinhardt. 2020.
\newblock \href {https://arxiv.org/abs/2009.03300} {Measuring massive multitask language understanding}.
\newblock \emph{arXiv preprint arXiv:2009.03300}.

\bibitem[{Hendrycks et~al.(2021)Hendrycks, Burns, Kadavath, Arora, Basart, Tang, Song, and Steinhardt}]{hendrycks2021measuring}
Dan Hendrycks, Collin Burns, Saurav Kadavath, Akul Arora, Steven Basart, Eric Tang, Dawn Song, and Jacob Steinhardt. 2021.
\newblock \href {https://arxiv.org/abs/2103.03874} {Measuring mathematical problem solving with the math dataset}.
\newblock In \emph{Thirty-fifth Conference on Neural Information Processing Systems Datasets and Benchmarks Track (Round 2)}.

\bibitem[{Hu et~al.(2023)Hu, Li, Zhao, Xie, Liu, Chen, Xie, and He}]{hu2023instructcoder}
Qisheng Hu, Kaixin Li, Xu~Zhao, Yuxi Xie, Tiedong Liu, Hui Chen, Qizhe Xie, and Junxian He. 2023.
\newblock \href {https://arxiv.org/abs/2310.20329} {Instructcoder: Empowering language models for code editing}.
\newblock \emph{arXiv preprint arXiv:2310.20329}.

\bibitem[{Jain et~al.(2024)Jain, Han, Gu, Li, Yan, Zhang, Wang, Solar-Lezama, Sen, and Stoica}]{jain2024livecodebench}
Naman Jain, King Han, Alex Gu, Wen-Ding Li, Fanjia Yan, Tianjun Zhang, Sida Wang, Armando Solar-Lezama, Koushik Sen, and Ion Stoica. 2024.
\newblock \href {https://arxiv.org/abs/2403.07974} {Livecodebench: Holistic and contamination free evaluation of large language models for code}.
\newblock \emph{arXiv preprint arXiv:2403.07974}.

\bibitem[{Jiao et~al.(2024)Jiao, Qin, Liu, Chen, and Joty}]{jiao2024learning}
Fangkai Jiao, Chengwei Qin, Zhengyuan Liu, Nancy~F Chen, and Shafiq Joty. 2024.
\newblock \href {https://arxiv.org/abs/2402.00658} {Learning planning-based reasoning by trajectories collection and process reward synthesizing}.
\newblock \emph{arXiv preprint arXiv:2402.00658}.

\bibitem[{Kuo et~al.(2024)Kuo, Halfaker, Cheng, Kim, Wu, Wu, Holstein, and Zhu}]{kuo2024wikibench}
Tzu-Sheng Kuo, Aaron~Lee Halfaker, Zirui Cheng, Jiwoo Kim, Meng-Hsin Wu, Tongshuang Wu, Kenneth Holstein, and Haiyi Zhu. 2024.
\newblock \href {https://dl.acm.org/doi/full/10.1145/3613904.3642278} {Wikibench: Community-driven data curation for ai evaluation on wikipedia}.
\newblock In \emph{Proceedings of the CHI Conference on Human Factors in Computing Systems}, pages 1--24.

\bibitem[{Li et~al.(2024)Li, Chiang, Frick, Dunlap, Wu, Zhu, Gonzalez, and Stoica}]{li2024crowdsourced}
Tianle Li, Wei-Lin Chiang, Evan Frick, Lisa Dunlap, Tianhao Wu, Banghua Zhu, Joseph~E Gonzalez, and Ion Stoica. 2024.
\newblock \href {https://arxiv.org/abs/2406.11939} {From crowdsourced data to high-quality benchmarks: Arena-hard and benchbuilder pipeline}.
\newblock \emph{arXiv preprint arXiv:2406.11939}.

\bibitem[{Li et~al.(2023)Li, Zhang, Dubois, Taori, Gulrajani, Guestrin, Liang, and Hashimoto}]{alpaca_eval}
Xuechen Li, Tianyi Zhang, Yann Dubois, Rohan Taori, Ishaan Gulrajani, Carlos Guestrin, Percy Liang, and Tatsunori~B. Hashimoto. 2023.
\newblock Alpacaeval: An automatic evaluator of instruction-following models.
\newblock \url{https://github.com/tatsu-lab/alpaca_eval}.

\bibitem[{Li et~al.(2025)Li, Zhang, Zhang, Zhang, Liu, Yao, Xu, Zheng, Wang, Chen et~al.}]{li2025system}
Zhong-Zhi Li, Duzhen Zhang, Ming-Liang Zhang, Jiaxin Zhang, Zengyan Liu, Yuxuan Yao, Haotian Xu, Junhao Zheng, Pei-Jie Wang, Xiuyi Chen, et~al. 2025.
\newblock From system 1 to system 2: A survey of reasoning large language models.
\newblock \emph{arXiv preprint arXiv:2502.17419}.

\bibitem[{Liang et~al.(2024)Liang, Chen, Wang, Wu, Fu, Shi, Wu, and Ye}]{liang2024robust}
Xize Liang, Chao Chen, Jie Wang, Yue Wu, Zhihang Fu, Zhihao Shi, Feng Wu, and Jieping Ye. 2024.
\newblock \href {https://arxiv.org/abs/2404.04102} {Robust preference optimization with provable noise tolerance for llms}.
\newblock \emph{arXiv preprint arXiv:2404.04102}.

\bibitem[{Lightman et~al.(2023)Lightman, Kosaraju, Burda, Edwards, Baker, Lee, Leike, Schulman, Sutskever, and Cobbe}]{lightman2023let}
Hunter Lightman, Vineet Kosaraju, Yura Burda, Harri Edwards, Bowen Baker, Teddy Lee, Jan Leike, John Schulman, Ilya Sutskever, and Karl Cobbe. 2023.
\newblock \href {https://arxiv.org/abs/2305.20050} {Let's verify step by step}.
\newblock \emph{arXiv preprint arXiv:2305.20050}.

\bibitem[{Lin et~al.(2024)Lin, Deng, Chandu, Brahman, Ravichander, Pyatkin, Dziri, Bras, and Choi}]{lin2024wildbench}
Bill~Yuchen Lin, Yuntian Deng, Khyathi Chandu, Faeze Brahman, Abhilasha Ravichander, Valentina Pyatkin, Nouha Dziri, Ronan~Le Bras, and Yejin Choi. 2024.
\newblock \href {https://arxiv.org/abs/2406.04770} {Wildbench: Benchmarking llms with challenging tasks from real users in the wild}.
\newblock \emph{arXiv preprint arXiv:2406.04770}.

\bibitem[{Liu et~al.(2024)Liu, Sferrazza, and Abbeel}]{liuchain}
Hao Liu, Carmelo Sferrazza, and Pieter Abbeel. 2024.
\newblock \href {https://arxiv.org/abs/2302.02676} {Chain of hindsight aligns language models with feedback}.
\newblock In \emph{The Twelfth International Conference on Learning Representations}.

\bibitem[{Liu et~al.(2021)Liu, Cui, Liu, Huang, Wang, and Zhang}]{liu2021logiqa}
Jian Liu, Leyang Cui, Hanmeng Liu, Dandan Huang, Yile Wang, and Yue Zhang. 2021.
\newblock \href {https://www.ijcai.org/proceedings/2020/0501.pdf} {Logiqa: a challenge dataset for machine reading comprehension with logical reasoning}.
\newblock In \emph{Proceedings of the Twenty-Ninth International Conference on International Joint Conferences on Artificial Intelligence}, pages 3622--3628.

\bibitem[{Ma et~al.(2024)Ma, Zhao, Zhang, He, and Kong}]{ma2024non}
Chang Ma, Haiteng Zhao, Junlei Zhang, Junxian He, and Lingpeng Kong. 2024.
\newblock \href {https://arxiv.org/abs/2410.17195} {Non-myopic generation of language models for reasoning and planning}.
\newblock \emph{arXiv preprint arXiv:2410.17195}.

\bibitem[{Meng et~al.(2025)Meng, Xia, and Chen}]{meng2025simpo}
Yu~Meng, Mengzhou Xia, and Danqi Chen. 2025.
\newblock \href {https://proceedings.neurips.cc/paper_files/paper/2024/hash/e099c1c9699814af0be873a175361713-Abstract-Conference.html} {Simpo: Simple preference optimization with a reference-free reward}.
\newblock \emph{Advances in Neural Information Processing Systems}, 37:124198--124235.

\bibitem[{Prasad et~al.(2024)Prasad, Yuan, Pang, Xu, Fazel-Zarandi, Bansal, Sukhbaatar, Weston, and Yu}]{prasad2024self}
Archiki Prasad, Weizhe Yuan, Richard~Yuanzhe Pang, Jing Xu, Maryam Fazel-Zarandi, Mohit Bansal, Sainbayar Sukhbaatar, Jason Weston, and Jane Yu. 2024.
\newblock \href {https://arxiv.org/abs/2411.04109} {Self-consistency preference optimization}.
\newblock \emph{arXiv preprint arXiv:2411.04109}.

\bibitem[{Qwen(2024)}]{qwq-32b-preview}
Team Qwen. 2024.
\newblock \href {https://qwenlm.github.io/blog/qwq-32b-preview/} {Qwq: Reflect deeply on the boundaries of the unknown}.

\bibitem[{Rafailov et~al.(2024)Rafailov, Sharma, Mitchell, Manning, Ermon, and Finn}]{rafailov2024direct}
Rafael Rafailov, Archit Sharma, Eric Mitchell, Christopher~D Manning, Stefano Ermon, and Chelsea Finn. 2024.
\newblock \href {https://proceedings.neurips.cc/paper_files/paper/2023/hash/a85b405ed65c6477a4fe8302b5e06ce7-Abstract-Conference.html} {Direct preference optimization: Your language model is secretly a reward model}.
\newblock \emph{Advances in Neural Information Processing Systems}, 36.

\bibitem[{Rein et~al.(2023)Rein, Hou, Stickland, Petty, Pang, Dirani, Michael, and Bowman}]{rein2023gpqa}
David Rein, Betty~Li Hou, Asa~Cooper Stickland, Jackson Petty, Richard~Yuanzhe Pang, Julien Dirani, Julian Michael, and Samuel~R Bowman. 2023.
\newblock \href {https://arxiv.org/abs/2311.12022} {Gpqa: A graduate-level google-proof q\&a benchmark}.
\newblock \emph{arXiv preprint arXiv:2311.12022}.

\bibitem[{Shao et~al.(2024)Shao, Wang, Zhu, Xu, Song, Bi, Zhang, Zhang, Li, Wu et~al.}]{shao2024deepseekmath}
Zhihong Shao, Peiyi Wang, Qihao Zhu, Runxin Xu, Junxiao Song, Xiao Bi, Haowei Zhang, Mingchuan Zhang, YK~Li, Y~Wu, et~al. 2024.
\newblock \href {https://arxiv.org/abs/2402.03300} {Deepseekmath: Pushing the limits of mathematical reasoning in open language models}.
\newblock \emph{arXiv preprint arXiv:2402.03300}.

\bibitem[{Sun et~al.(2024{\natexlab{a}})Sun, Chen, Xu, Cheng, Ma, Yin, Wang, Han, Zhu, Yuan et~al.}]{sun2024survey}
Qiushi Sun, Zhirui Chen, Fangzhi Xu, Kanzhi Cheng, Chang Ma, Zhangyue Yin, Jianing Wang, Chengcheng Han, Renyu Zhu, Shuai Yuan, et~al. 2024{\natexlab{a}}.
\newblock \href {https://arxiv.org/abs/2403.14734} {A survey of neural code intelligence: Paradigms, advances and beyond}.
\newblock \emph{arXiv preprint arXiv:2403.14734}.

\bibitem[{Sun et~al.(2024{\natexlab{b}})Sun, Cheng, Ding, Jin, Wang, Xu, Wu, Jia, Chen, Liu et~al.}]{sun2024genesis}
Qiushi Sun, Kanzhi Cheng, Zichen Ding, Chuanyang Jin, Yian Wang, Fangzhi Xu, Zhenyu Wu, Chengyou Jia, Liheng Chen, Zhoumianze Liu, et~al. 2024{\natexlab{b}}.
\newblock \href {https://arxiv.org/abs/2412.19723} {Os-genesis: Automating gui agent trajectory construction via reverse task synthesis}.
\newblock \emph{arXiv preprint arXiv:2412.19723}.

\bibitem[{Team et~al.(2023)Team, Anil, Borgeaud, Alayrac, Yu, Soricut, Schalkwyk, Dai, Hauth, Millican et~al.}]{team2023gemini}
Gemini Team, Rohan Anil, Sebastian Borgeaud, Jean-Baptiste Alayrac, Jiahui Yu, Radu Soricut, Johan Schalkwyk, Andrew~M Dai, Anja Hauth, Katie Millican, et~al. 2023.
\newblock \href {https://arxiv.org/abs/2312.11805} {Gemini: a family of highly capable multimodal models}.
\newblock \emph{arXiv preprint arXiv:2312.11805}.

\bibitem[{Teknium(2023)}]{OpenHermes-2.5}
Teknium. 2023.
\newblock \href {https://huggingface.co/datasets/teknium/OpenHermes-2.5} {Openhermes 2.5: An open dataset of synthetic data for generalist llm assistants}.

\bibitem[{Toshniwal et~al.(2024)Toshniwal, Moshkov, Narenthiran, Gitman, Jia, and Gitman}]{toshniwal2024openmathinstruct}
Shubham Toshniwal, Ivan Moshkov, Sean Narenthiran, Daria Gitman, Fei Jia, and Igor Gitman. 2024.
\newblock \href {https://arxiv.org/abs/2402.10176} {Openmathinstruct-1: A 1.8 million math instruction tuning dataset}.
\newblock \emph{arXiv preprint arXiv:2402.10176}.

\bibitem[{Trung et~al.(2024)Trung, Zhang, Jie, Sun, Jin, and Li}]{trung2024reft}
Luong Trung, Xinbo Zhang, Zhanming Jie, Peng Sun, Xiaoran Jin, and Hang Li. 2024.
\newblock \href {https://arxiv.org/abs/2401.08967} {Reft: Reasoning with reinforced fine-tuning}.
\newblock In \emph{Proceedings of the 62nd Annual Meeting of the Association for Computational Linguistics (Volume 1: Long Papers)}, pages 7601--7614.

\bibitem[{Uesato et~al.(2022)Uesato, Kushman, Kumar, Song, Siegel, Wang, Creswell, Irving, and Higgins}]{uesato2022solving}
Jonathan Uesato, Nate Kushman, Ramana Kumar, H~Francis Song, Noah~Yamamoto Siegel, Lisa Wang, Antonia Creswell, Geoffrey Irving, and Irina Higgins. 2022.
\newblock \href {https://mathai2022.github.io/papers/26.pdf} {Solving math word problems with process-based and outcome-based feedback}.

\bibitem[{Veeraboina(2023)}]{aime_1983_2024}
Hemish Veeraboina. 2023.
\newblock \href {https://www.kaggle.com/datasets/hemishveeraboina/aime-problem-set-1983-2024} {Aime problem set 1983-2024}.

\bibitem[{Wang et~al.(2024{\natexlab{a}})Wang, Li, Shao, Xu, Dai, Li, Chen, Wu, and Sui}]{wang2024math}
Peiyi Wang, Lei Li, Zhihong Shao, Runxin Xu, Damai Dai, Yifei Li, Deli Chen, Yu~Wu, and Zhifang Sui. 2024{\natexlab{a}}.
\newblock \href {https://aclanthology.org/2024.acl-long.510/} {Math-shepherd: Verify and reinforce llms step-by-step without human annotations}.
\newblock In \emph{Proceedings of the 62nd Annual Meeting of the Association for ComputatDo NOT Think That Much for 2+3=? On the Overthinking of o1-Like LLMsional Linguistics (Volume 1: Long Papers)}, pages 9426--9439.

\bibitem[{Wang et~al.(2024{\natexlab{b}})Wang, Ma, Zhang, Ni, Chandra, Guo, Ren, Arulraj, He, Jiang et~al.}]{wang2024mmlu}
Yubo Wang, Xueguang Ma, Ge~Zhang, Yuansheng Ni, Abhranil Chandra, Shiguang Guo, Weiming Ren, Aaran Arulraj, Xuan He, Ziyan Jiang, et~al. 2024{\natexlab{b}}.
\newblock \href {https://arxiv.org/abs/2406.01574} {Mmlu-pro: A more robust and challenging multi-task language understanding benchmark}.
\newblock \emph{arXiv preprint arXiv:2406.01574}.

\bibitem[{Wang et~al.(2025)Wang, Yue, and Chen}]{wang2025critique}
Yubo Wang, Xiang Yue, and Wenhu Chen. 2025.
\newblock \href {https://arxiv.org/abs/2501.17703} {Critique fine-tuning: Learning to critique is more effective than learning to imitate}.
\newblock \emph{arXiv preprint arXiv:2501.17703}.

\bibitem[{Xu et~al.(2024{\natexlab{a}})Xu, Sun, Cheng, Liu, Qiao, and Wu}]{xu2024interactive}
Fangzhi Xu, Qiushi Sun, Kanzhi Cheng, Jun Liu, Yu~Qiao, and Zhiyong Wu. 2024{\natexlab{a}}.
\newblock \href {https://arxiv.org/abs/2406.11736} {Interactive evolution: A neural-symbolic self-training framework for large language models}.
\newblock \emph{arXiv preprint arXiv:2406.11736}.

\bibitem[{Xu et~al.(2024{\natexlab{b}})Xu, Wu, Sun, Ren, Yuan, Yuan, Lin, Qiao, and Liu}]{xu2023symbol}
Fangzhi Xu, Zhiyong Wu, Qiushi Sun, Siyu Ren, Fei Yuan, Shuai Yuan, Qika Lin, Yu~Qiao, and Jun Liu. 2024{\natexlab{b}}.
\newblock \href {https://aclanthology.org/2024.acl-long.707/} {Symbol-llm: Towards foundational symbol-centric interface for large language models}.
\newblock In \emph{Proceedings of the 62nd Annual Meeting of the Association for Computational Linguistics (Volume 1: Long Papers)}, pages 13091--13116.

\bibitem[{Xu et~al.(2025{\natexlab{a}})Xu, Yan, Ma, Zhao, Liu, Lin, and Wu}]{xu2025phi}
Fangzhi Xu, Hang Yan, Chang Ma, Haiteng Zhao, Jun Liu, Qika Lin, and Zhiyong Wu. 2025{\natexlab{a}}.
\newblock \href {https://arxiv.org/abs/2503.13288} {$\phi$-decoding: Adaptive foresight sampling for balanced inference-time exploration and exploitation}.
\newblock \emph{arXiv preprint arXiv:2503.13288}.

\bibitem[{Xu et~al.(2025{\natexlab{b}})Xu, Wu, Wang, Li, Zheng, Chen, Hu, Kang, Ji, Zhang et~al.}]{xu2025redstar}
Haotian Xu, Xing Wu, Weinong Wang, Zhongzhi Li, Da~Zheng, Boyuan Chen, Yi~Hu, Shijia Kang, Jiaming Ji, Yingying Zhang, et~al. 2025{\natexlab{b}}.
\newblock \href {https://arxiv.org/abs/2501.11284} {Redstar: Does scaling long-cot data unlock better slow-reasoning systems?}
\newblock \emph{arXiv preprint arXiv:2501.11284}.

\bibitem[{Xu et~al.(2024{\natexlab{c}})Xu, Jiang, Niu, Deng, Poovendran, Choi, and Lin}]{xu2024magpie}
Zhangchen Xu, Fengqing Jiang, Luyao Niu, Yuntian Deng, Radha Poovendran, Yejin Choi, and Bill~Yuchen Lin. 2024{\natexlab{c}}.
\newblock \href {http://arxiv.org/abs/2406.08464} {Magpie: Alignment data synthesis from scratch by prompting aligned llms with nothing}.

\bibitem[{Yang et~al.(2024)Yang, Yang, Zhang, Hui, Zheng, Yu, Li, Liu, Huang, Wei et~al.}]{yang2024qwen2}
An~Yang, Baosong Yang, Beichen Zhang, Binyuan Hui, Bo~Zheng, Bowen Yu, Chengyuan Li, Dayiheng Liu, Fei Huang, Haoran Wei, et~al. 2024.
\newblock \href {https://arxiv.org/abs/2412.15115} {Qwen2. 5 technical report}.
\newblock \emph{arXiv preprint arXiv:2412.15115}.

\bibitem[{Yu et~al.(2020)Yu, Jiang, Dong, and Feng}]{yureclor}
Weihao Yu, Zihang Jiang, Yanfei Dong, and Jiashi Feng. 2020.
\newblock \href {https://arxiv.org/abs/2002.04326} {Reclor: A reading comprehension dataset requiring logical reasoning}.
\newblock In \emph{International Conference on Learning Representations}.

\bibitem[{Yuan et~al.(2024)Yuan, Pang, Cho, Li, Sukhbaatar, Xu, and Weston}]{yuanself}
Weizhe Yuan, Richard~Yuanzhe Pang, Kyunghyun Cho, Xian Li, Sainbayar Sukhbaatar, Jing Xu, and Jason~E Weston. 2024.
\newblock \href {https://arxiv.org/abs/2401.10020} {Self-rewarding language models}.
\newblock In \emph{Forty-first International Conference on Machine Learning}.

\bibitem[{Yue et~al.(2024)Yue, Qu, Zhang, Fu, Huang, Sun, Su, and Chen}]{yuemammoth}
Xiang Yue, Xingwei Qu, Ge~Zhang, Yao Fu, Wenhao Huang, Huan Sun, Yu~Su, and Wenhu Chen. 2024.
\newblock \href {https://arxiv.org/abs/2309.05653} {Mammoth: Building math generalist models through hybrid instruction tuning}.
\newblock In \emph{The Twelfth International Conference on Learning Representations}.

\bibitem[{Zelikman et~al.(2022)Zelikman, Wu, Mu, and Goodman}]{zelikman2022star}
Eric Zelikman, Yuhuai Wu, Jesse Mu, and Noah Goodman. 2022.
\newblock \href {https://proceedings.neurips.cc/paper_files/paper/2022/hash/639a9a172c044fbb64175b5fad42e9a5-Abstract-Conference.html} {Star: Bootstrapping reasoning with reasoning}.
\newblock \emph{Advances in Neural Information Processing Systems}, 35:15476--15488.

\bibitem[{Zeng et~al.(2024)Zeng, Huang, Zhao, Wang, Shan, and He}]{zeng2024b}
Weihao Zeng, Yuzhen Huang, Lulu Zhao, Yijun Wang, Zifei Shan, and Junxian He. 2024.
\newblock \href {https://arxiv.org/abs/2412.17256} {B-star: Monitoring and balancing exploration and exploitation in self-taught reasoners}.
\newblock \emph{arXiv preprint arXiv:2412.17256}.

\bibitem[{Zhang et~al.(2024)Zhang, Du, Pang, Liu, Gao, and Lin}]{zhang2024chain}
Xuan Zhang, Chao Du, Tianyu Pang, Qian Liu, Wei Gao, and Min Lin. 2024.
\newblock \href {https://arxiv.org/abs/2406.09136} {Chain of preference optimization: Improving chain-of-thought reasoning in llms}.
\newblock \emph{arXiv preprint arXiv:2406.09136}.

\end{thebibliography}
\bibliographystyle{acl_natbib}

\clearpage
\newpage
\appendix

\section{Gradient Analysis of ACO}
\label{app:gradient_derivation}

\paragraph{Derivation of the Gradient.}

The complete form of ACO loss function is written as:

% \begin{equation*}
%     \mathcal{L}_\textrm{ACO} = - \mathbb{E}_{(x,T^w,T^l,\alpha)} \mathrm{log} \sigma (\phi(x,T^w) - \phi_\alpha(x,T^l) )
% \end{equation*}
\begin{equation*}
\begin{split}
    \mathcal{L}_\textrm{ACO} = - \mathbb{E}_{(x,T^w,T^l) \sim \mathcal{D}} \mathrm{log} \sigma \left[ \beta \log \frac{\pi_\theta(T^w|x)}{\pi_\textrm{ref}(T^w|x)} \right. \\
    - \left. \beta \mathrm{clip} \left( \exp \frac{- (A_{l}-A_{w})}{\alpha}, 1 \right) \log \frac{\pi_\theta(T^l|x)}{\pi_\textrm{ref}(T^l|x)} \right]
\end{split}
\end{equation*}

To simplify the formulation, we denote the advantage calibration term as $w(x,A)$:
% \begin{equation*}
%     \phi_\alpha(x,T^l) = \beta w(x,A) \log \frac{\pi_\theta(T^l|x)}{\pi_\textrm{ref}(T^l|x)}
% \end{equation*}
\begin{equation*}
    w(x,A) = \mathrm{clip} \left( \exp \frac{- (A_{l}-A_{w})}{\alpha}, 1 \right) 
\end{equation*}
% \hangyan{maybe you shouldn't include $\beta$ in the above equation, because the following equation already has the term $\beta w(x,A)$}
Our objective is to derive the gradient $\nabla_\theta \mathcal{L}_\textrm{ACO}$. 
At the beginning, we define:

\begin{equation*}
    z = \beta \log \frac{\pi_\theta(T^w|x)}{\pi_\textrm{ref}(T^w|x)} - \beta w(x,A) \log \frac{\pi_\theta(T^l|x)}{\pi_\textrm{ref}(T^l|x)}
\end{equation*}

The gradient of the ACO loss can be represented in the form of:
\begin{equation*}
    \begin{aligned}
        \nabla_\theta \mathcal{L}_\textrm{ACO} &= - \mathbb{E}_{(x,T^w,T^l)} \nabla_\theta \log \sigma(z) \\
        &= - \mathbb{E}_{(x,T^w,T^l)} \frac{\sigma^{'}(z)}{\sigma(z)} \nabla_\theta z
    \end{aligned}
\end{equation*}

% \begin{equation*}
%     \textrm{where}, \nabla_\theta \log \sigma(z) = \frac{\sigma^{'}(z)}{\sigma(z)} \nabla_\theta z
% \end{equation*}

Using the unique characteristics of sigmoid function: $\sigma^{'}(x) = \sigma(x) (1- \sigma (x))$ and $\sigma(-x) = 1-\sigma(x)$, the gradient of ACO becomes:
% \hangyan{I think you didn't use the equation  $\sigma(-x) = 1-\sigma(x)$}
\begin{equation}
\label{eq_1}
    \nabla_\theta \mathcal{L}_\textrm{ACO} = - \mathbb{E}_{(x,T^w,T^l)} (1-\sigma(z)) \nabla_\theta z
\end{equation}

Now we need to compute $\nabla_\theta z$. Since $w(x,A)$ is independent of $\theta$, the gradient of $z$ reduces to:
\begin{equation*}
    \begin{aligned}
        \nabla_\theta z &= \beta \nabla_\theta \log \frac{\pi_\theta(T^w|x)}{\pi_\textrm{ref}(T^w|x)} \\
        &- \beta w(x,A) \nabla_\theta \log \frac{\pi_\theta(T^l|x)}{\pi_\textrm{ref}(T^l|x)}
    \end{aligned}
\end{equation*}

The gradients of the log probability terms are:
\begin{equation*}
    \nabla_\theta \log \frac{\pi_\theta(T^w|x)}{\pi_\textrm{ref}(T^w|x)} = \nabla_\theta \log \pi_\theta(T^w|x)
\end{equation*}
\begin{equation*}
    \nabla_\theta \log \frac{\pi_\theta(T^l|x)}{\pi_\textrm{ref}(T^l|x)} = \nabla_\theta \log \pi_\theta(T^l|x)
\end{equation*}

Replacing them into the gradient of $z$, we have:
\begin{equation}
\label{eq_2}
\begin{aligned}
    \nabla_\theta z &= \beta \nabla_\theta \log \pi_\theta(T^w|x) \\
    &- \beta w(x,A) \nabla_\theta \log \pi_\theta(T^l|x)
\end{aligned}
\end{equation}

Substituting Eq.~\ref{eq_2} into Eq.~\ref{eq_1},
the gradient of ACO is derived as:
\begin{equation}
\label{eq_3}
    \begin{split}
    \nabla_\theta \mathcal{L}_\textrm{ACO} = - \mathbb{E}_{(x,T^w,T^l)} \underbrace{(1-\sigma(z))}_{\text{scale}} \cdot \\ \underbrace{[\beta \nabla_\theta \log \pi_\theta(T^w|x) - \beta w(x,A) \nabla_\theta \log \pi_\theta(T^l|x)]}_{\text{postive and negative gradients}}
    \end{split}
\end{equation}

In this equation, $1-\sigma(z)$ controls the scale of the gradients. Putting $z$ in the formula, then we can obtain the following format:
\begin{equation}
\label{eq_4}
    \begin{aligned}
    1-\sigma(z)  &= \beta w(x,A) \log \frac{\pi_\theta(T^l|x)}{\pi_\textrm{ref}(T^l|x)} \\
    &- \beta \log \frac{\pi_\theta(T^w|x)}{\pi_\textrm{ref}(T^w|x)}
    \end{aligned}
\end{equation}

When the negative trajectory $T^l$ brings more advantages (i.e., higher $A$-value), $w(x,A)$ would decrease in value according to the illustration in Fig.~\ref{aco_loss}.
Then, the gradient scale $1-\sigma(z)$ drops, offering less optimization to the corresponding training data pairs.
It aligns with our initial motivation of calibrating with advantage values.

\paragraph{Relationship with Other Reinforced Loss Functions.}
With the derived gradient formulation, we discuss the relationship between ACO and other representative reinforced loss function~\cite{furuta2025geometric}.
DPO, c-DPO, and ROPO are included, where the latter two losses are specifically designed for the robust optimization.

\textbf{DPO.}
The formulation of DPO loss is:
\begin{equation*}
    \begin{aligned}
    \mathcal{L}_\textrm{DPO} &= - \log \sigma \left( \beta \log \frac{\pi_\theta(T^w|x)}{\pi_\textrm{ref}(T^w|x)} \right. \\ 
    & \left. - \beta \log \frac{\pi_\theta(T^l|x)}{\pi_\textrm{ref}(T^l|x)} \right)
    \end{aligned}
\end{equation*}
The gradient of DPO $\nabla_\theta \mathcal{L}_\textrm{DPO}$ can be represented as the same formulation of Eq.~\ref{eq_3} and ~\ref{eq_4}, where the gradient scale is $1-\sigma(z)$.
The DPO loss can be regarded as the special case of ACO loss, with $\alpha \rightarrow +\infty$.

\textbf{c-DPO.}
The c-DPO loss is designed to apply the label smoothing technique to alleviate the noise. 
Its formulation is:
\begin{equation*}
    \begin{aligned}
        &  \mathcal{L}_\textrm{c-DPO} =  \\
        & -\epsilon \log \sigma \left( \beta \log \frac{\pi_\theta(T^w|x)}{\pi_\textrm{ref}(T^w|x)} - \beta \log \frac{\pi_\theta(T^l|x)}{\pi_\textrm{ref}(T^l|x)} \right) \\
        & - (1-\epsilon) \log \sigma \left( \beta \log \frac{\pi_\theta(T^w|x)}{\pi_\textrm{ref}(T^w|x)} - \beta \log \frac{\pi_\theta(T^l|x)}{\pi_\textrm{ref}(T^l|x)} \right)
    \end{aligned}
\end{equation*}
% \hangyan{the above equation has some trouble in formatting}
The gradient of c-DPO $\nabla_\theta \mathcal{L}_\textrm{c-DPO}$ has the scale of $(1-\epsilon) - \sigma(z)$ with $\alpha \rightarrow +\infty$.
Compared with our ACO loss, c-DPO offers static and equal calibration for each data pair with the use of $\epsilon$, while ACO loss provides an adaptive solution.

\textbf{ROPO.}
The ROPO loss specially proposed for the robust optimization:
\begin{equation*}
    \begin{aligned}
        & \mathcal{L}_\textrm{ROPO} = -\gamma \sigma \left( \beta \frac{\pi_\theta(T^l|x)}{\pi_\textrm{ref}(T^l|x)} \sigma \frac{\pi_\theta(T^w|x)}{\pi_\textrm{ref}(T^w|x)} \right) \\
        &  -\eta \log \sigma \left( \beta \frac{\pi_\theta(T^w|x)}{\pi_\textrm{ref}(T^w|x)} \sigma \frac{\pi_\theta(T^l|x)}{\pi_\textrm{ref}(T^l|x)} \right) \\
    \end{aligned}
\end{equation*}
It has the similar gradient formulation with ACO loss when $\alpha \rightarrow +\infty$.
Its gradient scale can be derived as $(\gamma-\eta \sigma(z))(1-\sigma(z))$.
The main difference is ROPO provides the robust calibration with the positive and negative gradients, while ACO loss obtains
the advantage during the sampling process.

\section{Implementation Details}

% \subsection{Training Corpus and Evaluation Tasks}
% \label{app:train_test}

% \begin{table}[h]
% \centering
% \footnotesize
% \resizebox{\linewidth}{!}{
% \begin{tabular}{l|cc}
%     \toprule
%     \textbf{Training Corpus} &\textbf{\# Instructions} &\textbf{Sources} \\
%     \midrule
%     \multirow{1}{*}{Magpie} &25,000 &~\cite{xu2024magpie}  \\
%     \multirow{1}{*}{OpenHermes-2.5} &32,000 &~\cite{OpenHermes-2.5}  \\
%     \bottomrule
% \end{tabular}}
% \caption{Details of general-domain training corpus.}
% \label{appendix_train_data}
% \end{table}

% \begin{table*}[h]
% \centering
% \footnotesize
% \resizebox{\linewidth}{!}{
% \begin{tabular}{l|cccc}
%     \toprule
%     \textbf{Domains} &\textbf{Task name} &\textbf{\#Test Samples} &\textbf{\# Shot} &\textbf{Sources} \\
%     \midrule
%     \multirow{3}{*}{Math Reasoning} & GSM8K &1,319 &8 &\cite{cobbe2021training} \\
%     &MATH &4,001 &4 &\cite{hendrycks2021measuring} \\
%     &GPQA-Main &448 &0 &\cite{rein2023gpqa} \\
%     \midrule
%     \multirow{2}{*}{Logical Reasoning} & ReClor &500 &0 &\cite{yureclor} \\
%     & LogiQA &651 &0 &\cite{liu2021logiqa}\\
%     \midrule
%     \multirow{2}{*}{General Reasoning} & StrategyQA &2,290 &0 &\cite{geva2021did} \\
%     & ARC-Challenge &1,172 &0 &\citet{clark2018think}\\
%     \bottomrule
% \end{tabular}}
% \caption{Details of test tasks and benchmarks.}
% \label{appendix_task}
% \end{table*}

\subsection{Baselines}
\label{app:baselines}

\paragraph{SFT}
We finetune the LLM given the input query ($x$) and the labeled response ($a$).

\paragraph{SPIN~\cite{chenself}}
It iteratively refines the model-generated response against the labeled ones with an objective similar to DPO.

\paragraph{STaR~\cite{zelikman2022star}}
It continuously bootstraps from the self-constructed response through finetuning.

\paragraph{CoH~\cite{liuchain}}
It obtains both positive and negative responses via self-prompting and optimizes LLM with DPO loss function.

\paragraph{Self-Rewarding~\cite{yuanself}}
It leverages the LLM itself as a judge to label the self-generated responses (1-5 scores), then the LLM is optimized with DPO loss on the constructed preference pairs.

\paragraph{ScPO~\cite{prasad2024self}}
This approach generates multiple trajectories and labels the preference with self-consistency.
To address the open-ended generation scenarios, we modify the implementation of self-consistency with the cluster strategy.

\subsection{Setup}
\label{app:setup}

\paragraph{Sampling.}
The foresight sampling process is supported by 32*A100 GPUs of 80GB VRAM, and it is accelerated by the vLLM engine.
To ensure the diversity of sampling, we set the generation temperature to 0.6.
The step beam size $M$ is set to 2, the rollout number on each beam $N$ is set to 4, and the number of foresight steps $K$ is 4.

\paragraph{Training.}
The optimization of LLM is implemented with 8*A100 GPUs of 80GB VRAM, supported by Deepspeed Zero3 and FlashAttention2.
The total training batch size is set to 128, and the learning rate is 5e-7.
The hyper-parameter $\alpha$ in the ACO loss is set to 1.
Based on the configuration of sampling, we keep 4 training pairs for each query (collect one at each timestamp).
Therefore, the size of the training datasets is 100,000 and 128,000 for Magpie and OpenHermes-2.5 respectively.

\paragraph{Inference.}
The inference is supported by the vLLM engine.
We keep the default configuration of vLLM with a temperature of 1.0.
For GSM8K and MATH benchmarks, we leverage the widely-used few-shot examples, utilizing 4-shot for GSM8K and 8-shot for MATH.
For other benchmarks, we evaluate under the zero-shot setting.

\section{Analysis of The Training Corpus}
\label{app:train}

\begin{figure*}[htb]
\large
\centering
\includegraphics[scale=0.52]{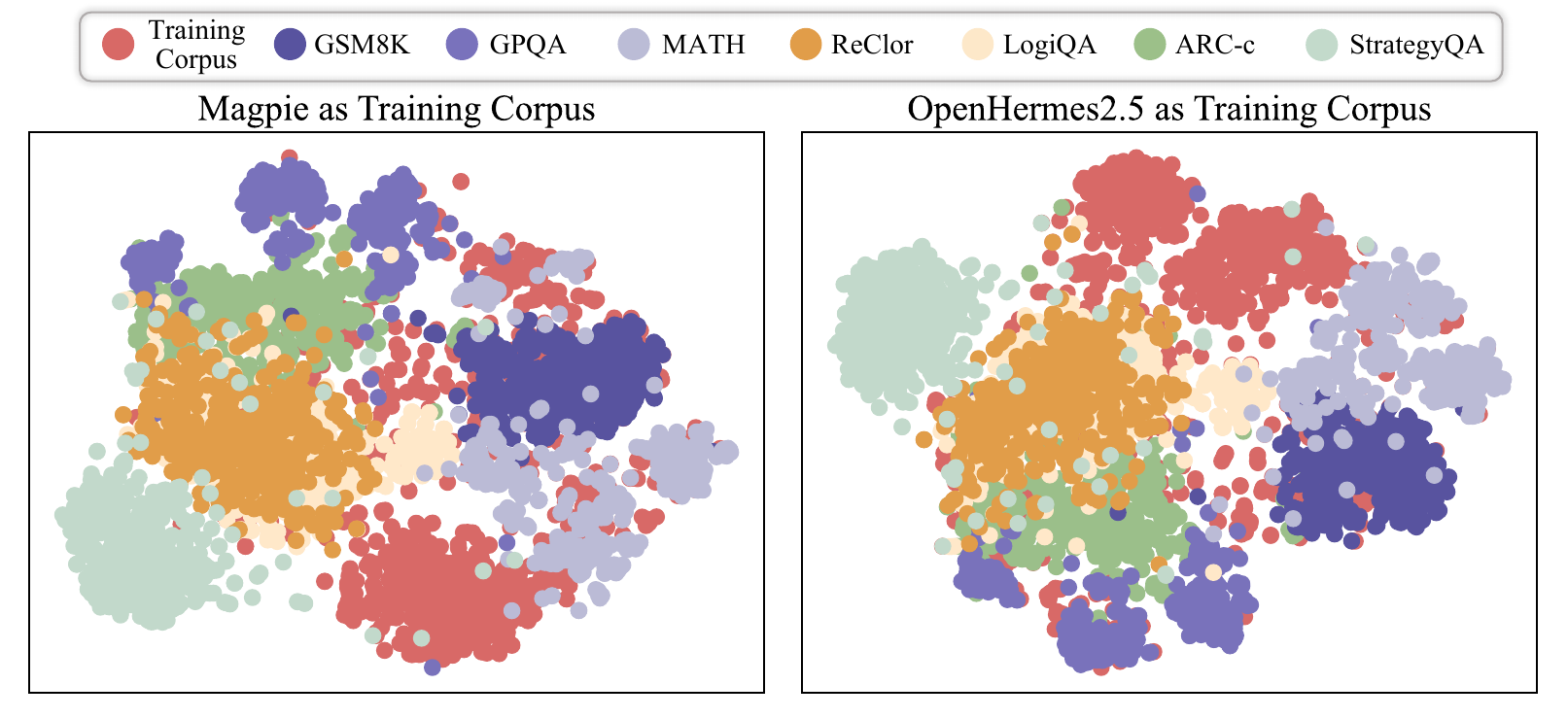}
\caption{Visualization of data distribution. The training corpus is marked in red color.}
\label{app:train_visualize}
% \vspace{-1em}
\end{figure*}

We visualize the data distribution difference between the training corpus and the evaluation tasks in Fig.~\ref{app:train_visualize}.
In the implementation, we utilize the sentence-embedding model ($\texttt{multi-qa-mpnet-base-dot-v1}$) to acquire the high-dimensional embeddings of the queries. Subsequently, we employ the t-SNE algorithm to visually represent these embeddings in a lower-dimensional space.
A set of 500 samples are randomly selected in each dataset.

To differentiate the data domains, we employ the following color scheme: \textbf{red} denotes the general training corpus, \textbf{purple} denotes mathematical datasets, \textbf{yellow} 
% \hangyan{maybe use orange is better}
denotes logical datasets, and \textbf{green} is used for other reasoning datasets.
% \hangyan{I think it's easier to read if all the four sentence use the same verb(e.g., red denotes XXX, purple denotes XXX ...... (it's just my opinion, it's up to you to decide)}
It is observed that the data distribution of the general training corpus is distinct and independent from that of other evaluation domains.
It supports our conclusion that \ours paves the way to self-improve LLM reasoning from unsupervised queries in the general domain.

Following the similarity analysis proposed in~\cite{xu2023symbol},
we also report the intra- and inter-class sample distances in Table~\ref{app:sim}.

\begin{table}[h]
\centering
\footnotesize
\resizebox{\linewidth}{!}{
\begin{tabular}{l|ccc}
    \toprule
    \textbf{Task} &\textbf{w/ Self} &\textbf{w/ Magpie} &\textbf{w/ OpenHermes} \\
    \midrule
    GSM8K &0.3967 &0.1343 &0.1256 \\
    MATH &0.1641 &0.0908 &0.0719 \\
    GPQA &0.1907 &0.0566 &0.0602 \\
    ReClor &0.2609 &0.0552 &0.0792 \\
    LogiQA &0.2330 &0.0732 &0.0930 \\
    StrategyQA &0.1186 &-0.0045 &0.0025 \\
    ARC-c &0.2276 &0.0673 &0.0843 \\
    \bottomrule
\end{tabular}}
\caption{Comparison between intra- and inter-class distance. \emph{w/ Self} denotes the distance within the task queries. \emph{w/ Magpie} means the distance between the target task with Magpie training corpus, while \emph{w/ OpenHermes} denotes the distance between the target task with OpenHermes2.5 training corpus.}
\label{app:sim}
\end{table}

The average similarity between the training corpus and the domain-specific downstream tasks, as depicted in the table, is notably low.
Among the tasks evaluated, the logical reasoning benchmarks (i.e., ReClor and LogiQA) exhibit the least resemblance to the training corpus, making them an ideal evaluation scenario for our approach.

\section{More Experiments on Coding Tasks}
Besides the natural language reasoning and understanding tasks,
it is also interesting to present the potential of \ours on coding-related benchmarks,
which is one of the key abilities of LLMs~\cite{sun2024survey}.
Table~\ref{app:code} reports the performances on MBPP~\cite{austin2021program} and LiveCodeBench~\cite{jain2024livecodebench}.

\begin{table}[h]
\centering
\footnotesize
\resizebox{\linewidth}{!}{
\begin{tabular}{l|cc}
    \toprule
    \textbf{Models} &\textbf{MBPP} &\textbf{LiveCodeBench} \\
    \midrule
    LLaMA3.1-8B-Instruct &69.65	&19.50 \\
    \textbf{\ours} [From Magpie] &71.60 &19.75\\
    \textbf{\ours} [From OpenHermes] &71.98 &21.25 \\
    \bottomrule
\end{tabular}}
\caption{Experiments on coding tasks.}
\label{app:code}
\end{table}

It is observed that self-training with \ours on the general data would also benefit the LLM coding abilities,
which involve strict formats and structured representations.

% \section{Analysis of $\alpha$ in ACO loss}

\end{document}